%% file: acl_latex.tex
\pdfoutput=1

\documentclass[11pt]{article}

\input{sections/header}
\title{MultiChartQA: Benchmarking Vision-Language Models\\on Multi-Chart Problems}

\author{
    Zifeng Zhu\thanks{Equal contribution.}\textsuperscript{1} \ \ \ \ \ \ \ \ Mengzhao Jia\footnotemark[1]\textsuperscript{2} \ \ \ \ \ \textbf{Zhihan Zhang\textsuperscript{2}} \ \ \ \textbf{Lang Li\textsuperscript{2}} \ \ \ \textbf{Meng Jiang\textsuperscript{2}} \\
\textsuperscript{1}Xi'an Jiaotong University \ \ \textsuperscript{2}University of Notre Dame \\
\texttt{zivenzhu@stu.xjtu.edu.cn} \ \ \texttt{\{mjia2,zzhang23,lli28,mjiang2\}@nd.edu} }

\begin{document}
\maketitle

\input{sections/abstract}

\input{sections/introduction}

\input{sections/related_work}

\input{sections/methodology}

\input{sections/experiment}

\input{sections/conclusion}

\nocite{auto-instruct}
\nocite{refaug}
\nocite{plug}

\bibliography{custom}

\newpage

\appendix



\input{appendices/question_categories}

\end{document}

%% file: sections/header.tex
\usepackage[preprint]{acl}

\newcommand{\ie}{\textit{i.e., }}
\newcommand{\eg}{\textit{e.g., }}

\usepackage{times}
\usepackage{latexsym}

\usepackage[T1]{fontenc}

\usepackage[utf8]{inputenc}

\usepackage{microtype}

\usepackage{inconsolata}

\usepackage{graphicx}

\usepackage{booktabs}
\usepackage{multirow}
\usepackage{subfigure}
\usepackage{cleveref}
\usepackage{enumitem}

%% file: sections/abstract.tex
\begin{abstract}

Multimodal Large Language Models (MLLMs) have demonstrated impressive abilities across various tasks, including visual question answering and chart comprehension, yet existing benchmarks for chart-related tasks fall short in capturing the complexity of real-world multi-chart scenarios. 
Current benchmarks primarily focus on single-chart tasks, neglecting the multi-hop reasoning required to extract and integrate information from multiple charts, which is essential in practical applications. To fill this gap, we introduce MultiChartQA, a benchmark that evaluates MLLMs' capabilities in four key areas: direct question answering, parallel question answering, comparative reasoning, and sequential reasoning. Our evaluation of a wide range of MLLMs reveals significant performance gaps compared to humans. These results highlight the challenges in multi-chart comprehension and the potential of MultiChartQA to drive advancements in this field. Our code and data are available at \url{https://github.com/Zivenzhu/Multi-chart-QA}.

\end{abstract}

%% file: sections/introduction.tex
\section{Introduction}

Multimodal Large Language Models (MLLMs) have received significant attention due to their impressive capabilities across a broad set of vision-language tasks, such as image captioning~\cite{agrawal2019nocaps, wang2023makes}, multimodal dialogue systems~\cite{saha2018towards, yin2024lamm}, and visual question answering (VQA)~\cite{schwenk2022okvqa, li2024seed,leopard}. Among their varied applications, the comprehension of visual chart data, such as bar and line charts, holds particular significance~\cite{chartllama,tinychart}. Such data is prevalent in real-world scenarios, including academic papers and analytical reports, which makes understanding and reasoning over these charts an essential skill for MLLMs. As a result, several efforts have been made in evaluating and improving model performance on chart-related tasks~\cite{masry2022chartqa,xu2023chartbench,chartgemma}. 

\begin{figure*}[t]
  \includegraphics[width=\textwidth] {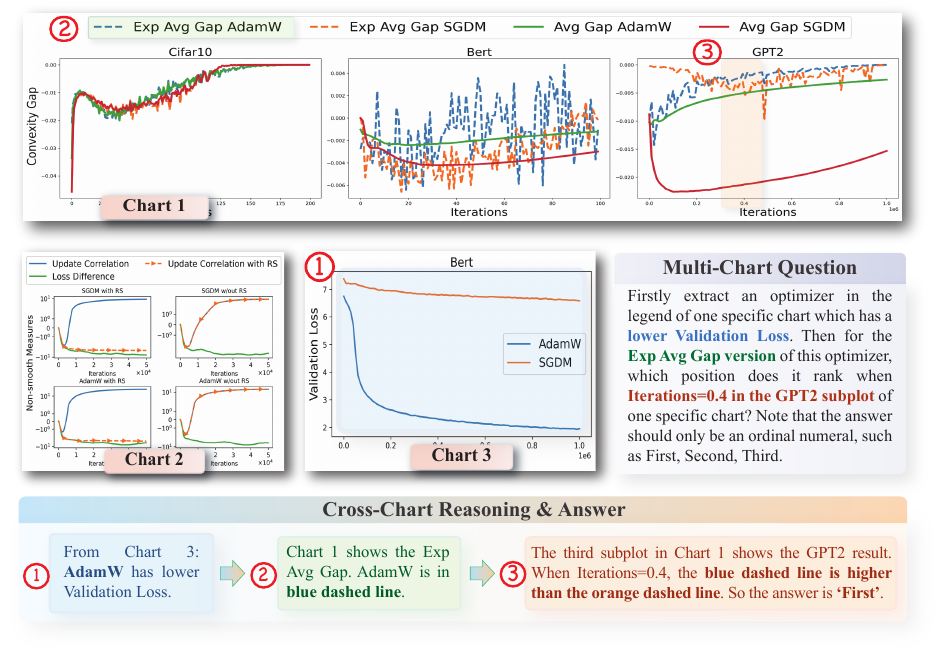}
  \centering
  \vspace{-1cm}
  \caption{An illustration of a multi-chart question is presented, asking for a comparison between the performance of two optimizers under specific conditions. The model is required to perform multi-hop reasoning across different charts to arrive at the correct answer. This scenario is frequently encountered in real-world applications. }
    \vspace{-0.5cm}
  \label{fig:intro}
\end{figure*}

Despite these efforts, existing chart understanding benchmarks are limited to single-chart scenarios \cite{kahou2017figureqa,kafle2018dvqa,methani2020plotqa,masry2022chartqa,xu2023chartbench,wang2024charxiv}. This narrow focus fails to cover evaluation settings involving multiple charts, which are common in real-world applications such as news articles and scientific documents. Such cases usually require MLLMs to integrate information from different charts and perform multi-hop reasoning to derive the answer to the question. For instance, Figure~\ref{fig:intro} demonstrates an example of a multi-chart question from a research paper. To determine whether the ``Exp Avg Gap'' for the optimizer ``AdamW'' exceeds that of ``SGDM'' under the GPT2 setting, information from multiple charts and sub-charts must be extracted and combined through a three-hop reasoning process to arrive at the correct answer. Such complex scenarios are hardly covered by single-chart benchmarks. In fact, solving these cases requires advanced capabilities of MLLMs, such as multi-image encoding, precise information localization, and multi-hop reasoning. The absence of complex multi-chart evaluation data hinders the further development of these advanced MLLM capabilities.

To bridge this gap, we introduce \textbf{MultiChartQA}, a newly collected benchmark designed to evaluate the capabilities of MLLMs on multi-chart scenarios. Our benchmark draws from a diverse set of public resources containing multi-chart articles, including Arxiv, OECD, Our World in Data, Pew Research Center, USAFacts, The World Economic Forum, Data Commons, International Energy Agency, GALLUP, and Gapminder. Using the charts collected from these sources, we design four different question types, each paired with multiple charts from the same article as input:
\begin{itemize}
[noitemsep,topsep=3pt,parsep=1pt,partopsep=0pt,leftmargin=0.4cm]
    \item \textit{Direct Question Answering} assesses whether the model can accurately identify the specific chart that contains the required information.
    \item \textit{Parallel Question Answering} increases the complexity by requiring the model to locate multiple charts and answer several information-seeking questions simultaneously.
    \item \textit{Comparative Reasoning} focuses on comparing different charts, requiring the model to grasp the distinctions and similarities among them.
    \item \textit{Sequential Reasoning} evaluates the model’s ability to perform multi-hop reasoning, interpreting charts in a stepwise manner.
\end{itemize}
Together, these four tasks provide a comprehensive evaluation of MLLMs' multi-chart processing abilities. All questions and answers in the benchmark are manually annotated to ensure their high quality.

We evaluate the performance of 20 mainstream MLLMs on MultiChartQA. The results reveal a significant limitation in MLLMs' ability to process multiple charts, with performance trailing behind humans by a large margin. Furthermore, there is a notable gap between the performance of proprietary models and their open-source counterparts. Further analysis conducted on our benchmark highlights the benefits of using chain-of-thought reasoning, as well as the models' reliance on chart references for information retrieval. An in-depth error analysis reveals that MLLMs still struggle with visual perception and multi-step reasoning, especially when color or spatial constraints are involved. We believe that MultiChartQA will illuminate the evaluation of multi-chart capabilities in MLLMs and foster advancements in this field in the near future.

\begin{figure*}[!t]
  \includegraphics[width=\textwidth]{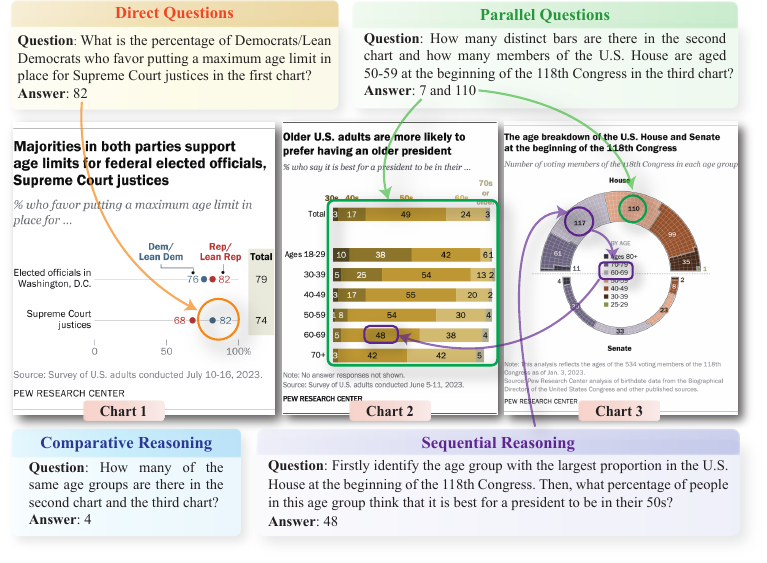}
  \vspace{-1cm}
  \caption{Multi-ChartQA contains four types of QA tasks, covering four crucial abilities for understanding and reasoning across multiple charts. 
  We highlight the key information location for answering each question with boxes and circles. The arrows represent the multi-step reasoning process across different charts.}
    \vspace{-0.5cm}
  \label{fig:method}
\end{figure*}

%% file: sections/related_work.tex
\section{Related Work}

\subsection{Chart Question Answering Benchmarks}
Many chart question-answering benchmarks have been developed over time. Early examples, such as FigureQA \cite{kahou2017figureqa}, DVQA \cite{kafle2018dvqa}, PlotQA \cite{methani2020plotqa}, ChartBench \cite{xu2023chartbench} and ChartX \cite{xia2024chartx}, primarily utilize Python packages to produce charts, resulting in limited diversity of chart types. More recent benchmarks, including ChartQA \cite{masry2022chartqa} and Charxiv \cite{wang2024charxiv}, collect real-world charts and adopt human annotation to ensure diversity in both charts and questions. However, these studies do not address multi-chart scenarios. 
Although MMC-Benchmark \cite{liu2023mmc} includes a small subset of multichart data, it comprises only 52 samples from a single source, which is insufficient to comprehensively evaluate models' performance on multichart tasks.

\subsection{Multi-image VQA Benchmarks}

Early research concentrated on evaluating single-image settings~\cite{schwenk2022okvqa,goyal2017making}. Recently, multi-image question-answering benchmarks like Mantis-Instruct~\cite{jiang2024mantis}, BLINK~\cite{fu2024blink}, and MUIRBENCH~\cite{wang2024muirbench} have been introduced. However, these benchmarks assess multi-image understanding in general domains—encompassing natural images, remote sensing, medical imagery, and others—but leave the chart domain unaddressed. Although ReMI~\cite{kazemi2024remi} includes some multi-chart scenarios, it uses synthetic charts and a single type of questions, which is insufficient for capturing the complex reasoning involved in practical multi-chart analyses.
In contrast, our benchmark focuses specifically on chart-domain evaluation. We design four types of questions aligned with real-world multi-chart scenarios, making MultiChartQA an effective evaluation tool for those developing chart-domain expert MLLMs.

\subsection{Chart Understanding}
Chart understanding is a vital task that involves interpreting chart content to perform functions such as data extraction~\cite{jung2017chartsense}, summarization~\cite{xu2018chart}, and question-answering~\cite{kafle2018dvqa,hoque2022chart,masry2022chartqa}. Recent advancements can be divided into two main categories: two-stage methods and end-to-end methods. Two-stage methods use specialized extraction modules to generate intermediary representations of chart information—such as tables—that serve as textual prompts for large language models~\cite{jia2024describe,liu2022matcha}. End-to-end methods strive to tackle chart reasoning challenges with unified models by integrating multimodal large language models to enhance understanding capabilities~\cite{chartllama,liu2023mmc,liu2024chartthinker,tinychart}.
While these end-to-end models have demonstrated improved performance, their training resources contain limited multi-chart samples, resulting in challenges in their ability to generalize to diverse or complex multi-chart tasks.

%% file: sections/methodology.tex
\section{MultiChartQA}

MultiChartQA is an extensive and demanding benchmark that features real-world charts and complex reasoning questions. To cover charts with various topics and a diverse range of styles, we curate chart sets from a variety of real-world sources (\S\ref{3.1: Chart Source}). Based on four core competencies in chart comprehension, we design four categories of question-answer pairs  (\S\ref{3.2: Question Design}). We detail the evaluation metrics in \S\ref{3.3: Our Evaluation Metric}. Ultimately, the MultiChartQA benchmark comprises 1,370 charts and 2,000 questions.

\begin{table}[]
    \centering
    \scalebox{0.9}{
    \begin{tabular}{lc}
    \hline
     \toprule
     \textbf{Statistic} & \textbf{Number} \\
     \midrule
      Total Questions & 2,000 \\
     \textbf{Category} & \\
      ~- Direct Question & 527 (26.35\%) \\
      ~- Parallel Questions & 617 (30.85\%) \\
      ~- Comparative Reasoning & 581 (29.05\%) \\
      ~- Sequential Reasoning  & 275 (13.75\%) \\
       \textbf{Type} & \\
       ~-Multiple-choice & 247 (12.35\%)\\
      ~- Open-ended & 1,753 (87.65\%)\\
     \midrule
     Unique charts & 1,370 \\
     Multi-chart sets  & 500 \\
     Average charts & 2.74 \\
      \textbf{Source} & 10 \\
     \midrule
     \textbf{Questions} &\\
     Unique questions & 1,646 \\
     Unique tokens & 4,341 \\
     Maximum/Average length & 162/36.4 \\
     \midrule 
     \textbf{Answers} & \\
     Unique answers & 1,109 \\
     Unique tokens & 1,289 \\
     Maximum/Average length & 55/3.1\\
     \bottomrule
     \hline
     \end{tabular}}
     
     \caption{Statistics of MultiChartQA. We analyze the characteristics of charts, questions, and answers. The unique tokens and lengths of questions and answers are measured using the GPT-4o tokenizer.}
     \label{tab:statistics}
     \vspace{-0.3cm}
\end{table}

\begin{figure}[t]
\centering
  \includegraphics[width=0.7\columnwidth]
  {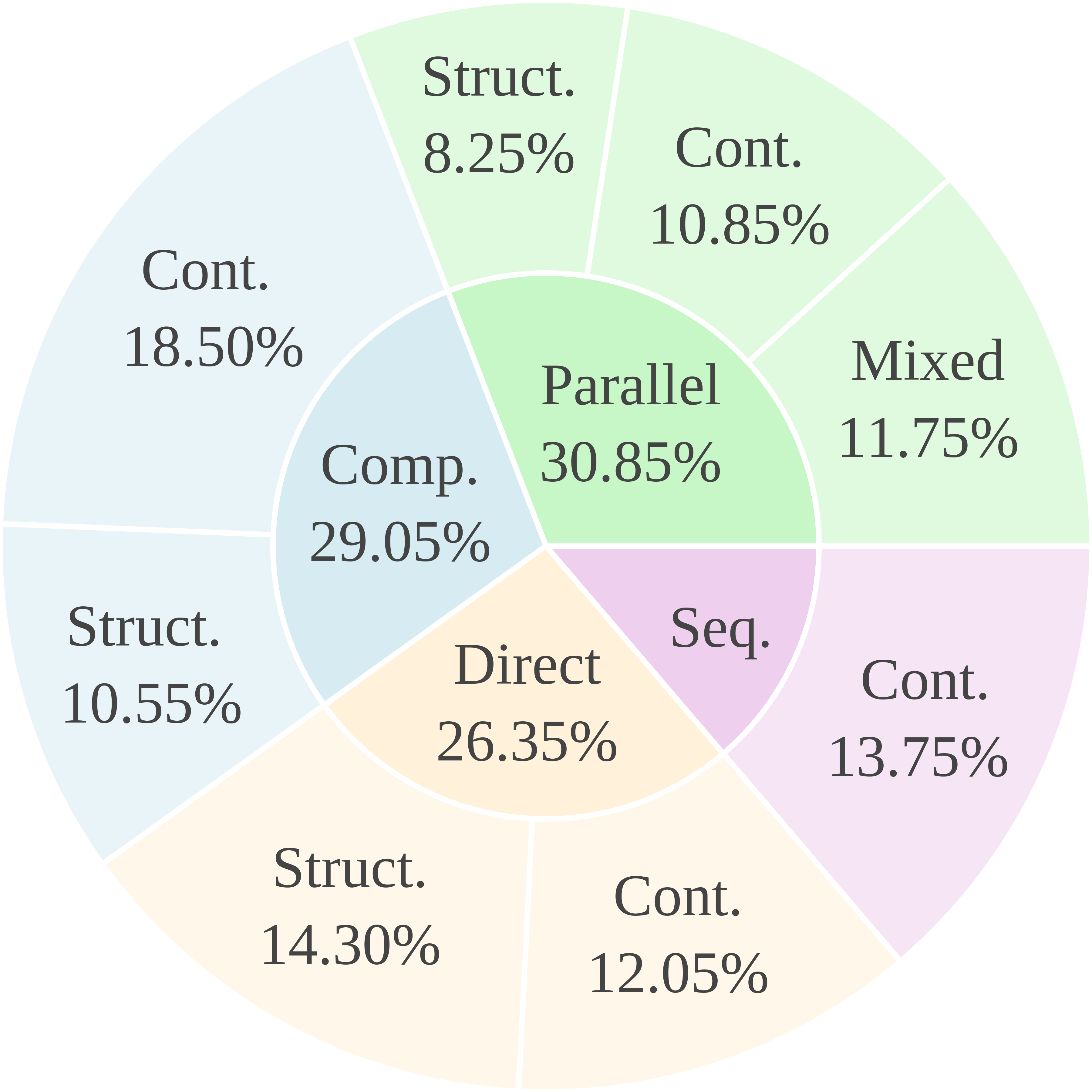}
  \caption{Detailed illustration of question categories. MultiChartQA features four distinct types of questions, varying in form, content, and difficulty. For brevity, the category names are abbreviated. Struct.: Structure, Comp.:Comparative, and Seq.: Sequential.}
  \label{fig:question_statistics}
    \vspace{-0.5cm}
\end{figure}

\subsection{Chart Collection}
\label{3.1: Chart Source}

For MultiChartQA, we aim to utilize chart sets that maintain semantic coherence and topic consistency. The charts within the same article or paper naturally fulfill this criterion. We therefore identify ten websites that provide such multi-chart content, namely ArXiv\footnote{\url{https://arxiv.org/}}, OECD\footnote{\url{https://www.oecd.org/}}, Our World in Data\footnote{\url{https://ourworldindata.org/}}, Pew Research Center\footnote{\url{https://www.pewresearch.org/}}, USAFacts\footnote{\url{https://usafacts.org/}}, The World Economic Forum\footnote{\url{https://www.weforum.org/}}, Data Commons\footnote{\url{https://datacommons.org/}}, International Energy Agency\footnote{\url{https://tinyurl.com/27ywfv6n}}, GALLUP\footnote{\url{https://www.gallup.com/}}, and Gapminder\footnote{\url{https://www.gapminder.org/tools/}}. We collect chart sets from these sources, covering diverse topics.

Next, we introduce the details of the collection of these sources. \textbf{ArXiv} is an open-access repository that hosts preprints of research papers across various scientific disciplines. We extracted images from 1,500 publications using the arXiv API\footnote{\url{https://info.arxiv.org/help/api/index.html}}. Since research papers often contain a variety of images beyond charts, such as natural images or diagrams, we employ an off-the-shelf MLLM, LLaVA-1.5-13B~\cite{liu2024improved}, in a zero-shot prompting approach to classify and filter out non-chart images. We consider multiple charts from the same paper as a chart set. To ensure the charts in one set have strong semantic correlations, we manually review the sets and retain those that share one or more elements, \eg same entities or categories. In total, we collect 129 sets of charts from Arxiv. 
\textbf{International Energy Agency} contains a variety of charts in different formats and on diverse topics. We manually select charts on this website and group those with the same topic into a chart set, resulting in the collection of 32 multi-chart sets. Similarly, we conduct manual chart collection from \textbf{Gapminder} website, generating a total of 4 chart sets. 
As to \textbf{OECD}, \textbf{Our World in Data}, \textbf{Pew Research Center}, \textbf{USAFacts}, \textbf{The World Economic Forum}, \textbf{Data Commons}, and \textbf{GALLUP}, we resort to python crawlers for chart downloading, followed by human double-check. We treat the charts from the same report as a semantically related chart set. In total, we collected 58, 58, 56, 50, 48, 44, and 21 multi-chart sets from the aforementioned sources, respectively.
Overall, 500 multi-chart sets are collected, each containing 2-3 semantically related charts. 
The proportion of each source is shown in Appendix~\ref{appendix:charts_source}. Some examples of chart sets from different sources can be found in Appendix~\ref{appendix:charts}.

\subsection{Question Design}
\label{3.2: Question Design}

We concentrate on four key capabilities in multi-chart tasks and design four types of questions, \ie direct questions, parallel questions, comparative reasoning, and sequential reasoning. Example questions of each category are illustrated in Figure~\ref{fig:method}. We manually annotate every question and answer to ensure correctness. A quality evaluation is conducted after the annotation process to double-check all questions and fix the inaccurate ones.

\paragraph{Direct Questions.}
The direct questions evaluate whether the model can accurately identify the relevant chart to answer questions accordingly. We present multiple charts and use terms like ``In the second chart'' to explicitly specify which chart the answer should pertain to. The design of the questions focuses on two types of chart information: \textbf{structural} information and \textbf{content} information. The former involves details about the chart's structure, such as the values on the axes, the layout of subplots, and the labels in the legend~\cite{wang2024charxiv}. These details are essential for comprehending the overall architecture of the chart and interpreting its organizational form. 
The latter focuses on the chart's content, such as extracting specific numbers or phrases based on the given clues. This type of task requires the model to understand the semantics of the chart content to perform precise reasoning and information analysis.

\paragraph{Parallel Questions.}
Unlike direct questions that ask the model to locate and focus on one specific chart, parallel questions examine the model's ability to answer independent questions that span multiple charts simultaneously. 
Such questions often arise in scenarios where information is gathered across semantically related charts. These questions are formed by querying information from multiple charts, with each parallel question containing several sub-questions, and each sub-answer is derived from a different chart. Similar to direct questions, we explicitly specify which chart the answer should come from. Like direct questions, parallel questions also include \textbf{structure} questions, \textbf{content} questions, as well as \textbf{mixed} questions that involve both.

\paragraph{Comparative Reasoning.}
Comparison questions assess the model's ability to analyze and compare information across multiple charts, requiring reasoning between them. For instance, in Figure 1, the question "How many same age groups are there in the second chart and the third chart?" requires the model to compare the age groups in both charts and identify the overlapping ones. Like the previous two question types, comparison questions are divided into two categories: those focus on \textbf{structure} and those focus on \textbf{content}.

\paragraph{Sequential Reasoning.}
Sequential reasoning involves complex multi-step reasoning questions with a temporal or logical sequence. To solve such problems, the model needs to track and analyze different aspects of an entity from the information dispersed in different charts. 
Specifically, these questions use a single entity as a clue and, through multi-hop reasoning, traverse several charts to arrive at the final answer. The design of these questions is inspired by the human process of reasoning across multiple charts, where one must follow an entity's various characteristics or attributes across charts. All sequential reasoning questions are \textbf{content-related}.

\begin{table*}[t!]
  \centering
  \scalebox{0.91}{
      \begin{tabular}{lrrrrrrrrrrrrrr@{}}
        \hline
        \toprule
        \multirow{3}{*}{\textbf{Model}} &
        \multirow{3}{*}{\textbf{Overall}} & \phantom{} &
        \multicolumn{2}{c}{\textbf{Direct}} & \phantom{} &
        \multicolumn{3}{c}{\textbf{Parallel}} & \phantom{} &
        \multicolumn{2}{c}{\textbf{Comparative}} & \phantom{} &
        \textbf{Seq.} & \phantom{} \\
        
        \cmidrule{4-5} \cmidrule{7-9} \cmidrule{11-12} \cmidrule{14-14}
        
        \phantom{} & \phantom{} & & \small \textbf{Struct.} & \small \textbf{Cont.} &  & \small \textbf{Struct.} & \small \textbf{Mixed} & \small \textbf{Cont.} & & \small \textbf{Struct.} & \small \textbf{Cont.} &  & \small \textbf{Cont.} \\

        \midrule
        
        Human  & \textbf{90.11} &  & \textbf{95.49} & \textbf{92.41} &   & \textbf{91.88} & \textbf{94.50} & \textbf{87.50} &   & \textbf{91.11} & \textbf{85.21} &   & \textbf{83.16}  \\

        Random (GPT-4o) & 13.96 &  & 5.24 & 6.22 &   & 10.61 & 11.77 & 15.67 &   & 5.21 & 31.35 &   & 15.64  \\

        \midrule
            \multicolumn{15}{c}{\textbf{Closed-source Models}} \\
        \midrule

        Claude-3.5-Sonnet  & \textbf{77.72} &  & \textbf{86.36} & \textbf{78.42} &   & \textbf{81.31} & 80.92 & \textbf{80.65} &   & 72.51 & \textbf{79.46} &   & \textbf{62.55}  \\

        GPT-4o  & 72.46 &  & 80.77 & 71.37 &   & 76.87 & \textbf{81.42} & 74.42 &   & \textbf{75.83} & 70.41 &   & 53.09  \\

        Gemini-1.5-Pro  & 66.83 &  & 77.97 & 71.78 &   & 73.84 & 73.69 & 62.21 &   & 63.03 & 65.81 &   & 48.73  \\

        GPT-4o-mini  & 59.27 &  & 72.38 & 60.58 &   & 61.62 & 63.48 & 56.91 &   & 49.76 & 64.59 &   & 41.45  \\
        
        \midrule
            \multicolumn{15}{c}{\textbf{Open-source General-purpose Models}} \\
        \midrule

        Qwen2-VL-72B & \textbf{65.67} &  & \textbf{81.47} & \textbf{72.61} &   & \textbf{60.61} & \textbf{68.87} & \textbf{64.98} &   & \textbf{52.13} & \textbf{73.11} &   & \textbf{44.36}  \\

        MiniCPM-V2.6 & 53.13 &  & 69.93 & 56.85 &   & 47.98 & 54.47 & 48.85 &   & 29.38 & 66.62 &   & 37.82  \\

        LLaVA-OV-72B & 53.09 &  & 67.48 & 62.24 &   & 52.63 & 62.34 & 55.30 &   & 28.91 & 58.24 &   & 32.36  \\

        InternVL2-26B & 50.54 &  & 65.03 & 57.26 &   & 46.57 & 55.32 & 52.30 &   & 33.18 & 55.00 &   & 33.82  \\

        Qwen2-VL-7B  & 47.03 &  & 64.69 & 61.83 &   & 40.00 & 48.16 & 50.00 &   & 16.59 & 60.27 &   & 22.18  \\
        
        InternVL-V1.5  & 46.42 &  & 60.14 & 56.02 &   & 38.99 & 48.79 & 46.77 &   & 27.01 & 55.68 &   & 28.36  \\

        LLaVA-OV-7B & 36.12 &  & 50.70 & 42.32 &   & 29.70 & 36.10 & 39.17 &   & 12.32 & 51.76 &   & 14.18  \\

        MiniCPM-V2.5  & 33.27 &  & 45.45 & 22.82 &   & 30.10 & 36.10 & 32.03 &   & 20.38 & 45.81 &   & 23.27  \\

        DeepSeek-VL-7B  & 29.42 &  & 34.27 & 32.78 &   & 23.23 & 27.87 & 28.57 &   & 13.27 & 43.11 &   & 21.09  \\

        Eagle-X5-13B & 28.07 &  & 34.97 & 38.17 &   & 20.30 & 26.95 & 25.81 &   & 12.80 & 37.97 &   & 17.82  \\
        
        LLaVA-V1.6-7B & 26.35 &  & 26.57 & 24.90 &   & 23.13 & 27.59 & 31.34 &   & 9.95 & 37.57 &   & 21.82  \\

        Idefics3-8B & 13.48 &  & 11.54 & 12.86 &   & 9.19 & 12.98 & 17.74 &   & 2.37 & 24.46 &   & 9.45  \\
        
        Idefics2-8B  & 11.87 &  & 7.34 & 14.11 &   & 8.89 & 10.92 & 11.06 &   & 6.64 & 22.43 &   & 7.64  \\

        \midrule
            \multicolumn{15}{c}{\textbf{Chart-Domain Specialized Models}} \\
        \midrule

        ChartGemma & \textbf{11.91} &  & \textbf{6.99} & \textbf{13.28} &   & 5.76 & 9.22 & \textbf{10.83} &   & \textbf{6.64} & \textbf{22.84} &   & \textbf{12.00}  \\

        TinyChart-3B  & 10.17 &  & \textbf{6.99} & 7.47 &   & \textbf{7.88} & \textbf{9.79} & 7.83 &   & 6.16 & 20.68 &   & 8.36  \\
        
        MatCha & 6.83 &  & 7.34 & 8.71 &   & 3.13 & 2.27 & 5.07 &   & 2.37 & 16.22 &   & 2.91  \\
        

        \bottomrule
        \hline
      \end{tabular}
    }
    \caption{Evaluation results on MultiChartQA. \textbf{Bold} values indicate the best performance within each category: closed-source or open-source models. Question categories and model names are abbreviated due to space limits. Struct.: Structure, Cont.: Content, Seq.: Sequential, LLaVA-OV-72B: LLaVA-OneVision-72B, LLaVA-OV-7B: LLaVA-OneVision-7B}
  \vspace{-2ex}
  \label{tab:main_result}
\end{table*}

\subsection{Evaluation Metric} 
\label{3.3: Our Evaluation Metric}
We add explicit instructions for each question to control the output format, enabling efficient extraction of answers. The model outputs are parsed and answers are extracted using scripts\footnote{\url{https://github.com/MMMU-Benchmark/MMMU}.}. We then evaluate numerical answers using relaxed accuracy~\cite{methani2020plotqa, masry2022chartqa} and non-numeric answers using exact match to determine whether the extracted answer aligns with the ground truth. For questions that consist of multiple sub-questions, \ie parallel questions, we calculate accuracy based on the proportion of correctly answered sub-questions, ensuring a fair and fine-grained evaluation.


%% file: sections/experiment.tex
\section{Experiments}

\begin{figure*}[t]
    \includegraphics[width=\textwidth]{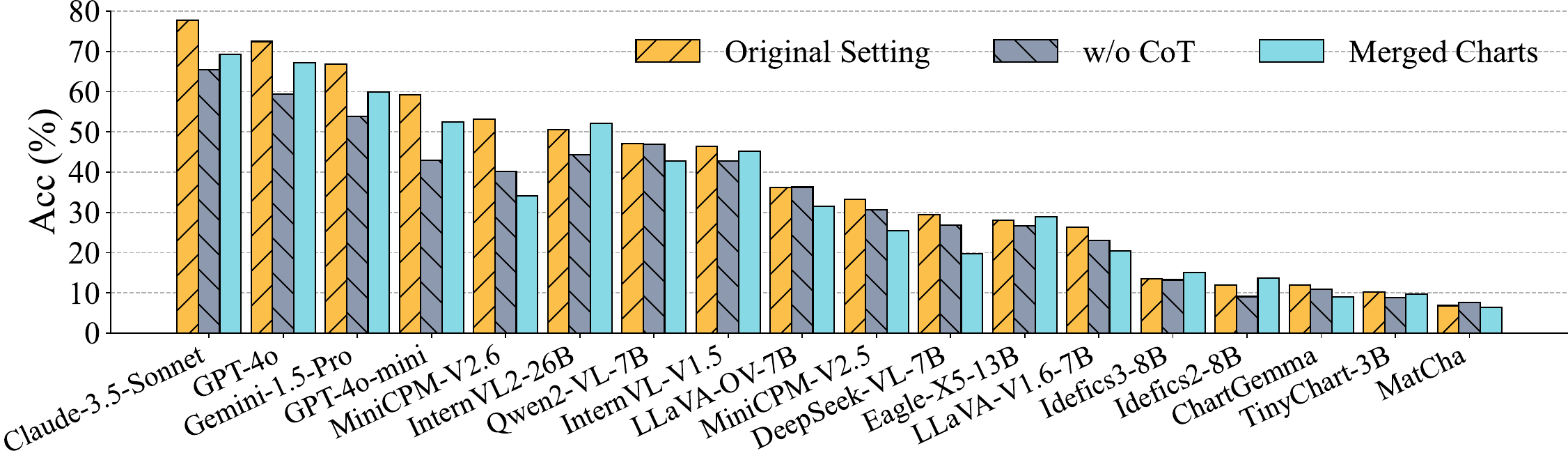}
    \vspace{-0.5cm}
    \caption{The accuracy of the 18 MLLMs is evaluated under three settings: the original setting, merged charts, and without Chain-of-Thought reasoning. Most models exhibit a decline in performance when processing merged charts or when answering without conducting CoT.}
    \label{fig:CoT_&_merged}
    \vspace{-0.5cm}
\end{figure*}
\subsection{Experimental Setup}

\paragraph{Evaluated MLLMs.} We evaluate 20 mainstream MLLMs on our benchmark, comprising 4 closed-source models, 13 open-source general-purpose models, and 3 chart-domain specialized models. The closed-source models include Claude-3.5-Sonnet~\cite{claude3.5-sonnet}, GPT-4o~\cite{gpt4o}, GPT-4o-mini~\cite{gpt4o}, and Gemini-1.5-Pro~\cite{reid2024gemini}. Among the open-source general-purpose models, we evaluate Qwen2-VL-72B~\cite{wang2024qwen2}, Qwen2-VL-7B~\cite{wang2024qwen2}, MiniCPM-V2.6~\cite{yao2024minicpm}, MiniCPM-V2.5~\cite{yao2024minicpm}, LLaVA-OneVision-72B~\cite{li2024llava}, LLaVA-OneVision-7B~\cite{li2024llava}, LLaVA-V1.6-7B~\cite{liu2024llava}, InternVL2-26B~\cite{chen2024far}, InternVL-V1.5~\cite{chen2024far}, DeepSeek-VL-7B~\cite{lu2024deepseek}, Eagle-X5-13B~\cite{shi2024eagle}, Idefics3-8B~\cite{laurenccon2024building}, and Idefics2-8B~\cite{laurenccon2024matters}. For the chart-model, we evaluate ChartGemma~\cite{masry2024chartgemma}, TinyChart-3B~\cite{zhang2024tinychart} and MatCha~\cite{liu2022matcha}. For reference, we include human performance and random choice baselines. For the former, we asked 11 graduate student volunteers with sufficient training to answer the questions. For the latter, following~\cite{wang2024charxiv}, we use GPT-4o to generate random guesses without access to the charts, with the exact prompt detailed in Appendix~\ref{appendix:prompt}.

\subsection{Experimental Results}
Table~\ref{tab:main_result} demonstrates the performance of 20 MLLMs on MultiChartQA, from which we derive the following observations.
\textbf{There is a notable disparity between closed-source and open-source MLLMs}. The best closed-source model, Claude-3.5-Sonnet, achieves an overall accuracy of 77.72\%. In contrast, most open-source models fall short in performance—except for one, namely Qwen2-VL-72B, which attains an impressive performance of 65.67\%. This accuracy gap underscores the effectiveness of our benchmark in distinguishing model capabilities through comprehensive questioning.


\textbf{Models perform poorly on sequential reasoning among four categories.} The sequential reasoning questions present significant challenges for MLLMs. For example, Claude-3.5-Sonnet reaches only 62.55\% accuracy on this task, with GPT-4o following at 53.09\%. The best open-source model, Qwen2-VL-72B, achieves only 44.36\%, showing a remarkable performance gap between models and humans.  

\textbf{Models exhibit performance degradation on parallel questions compared to direct ones.} Although the difficulty of each sub-question in parallel questions is comparable to that of direct questions, the model's performance drops significantly on parallel questions, especially for content-related ones. This indicates that, beyond the challenges posed by individual sub-questions, handling multiple charts simultaneously introduces additional difficulties for the model.

\textbf{Chart-domain specialized models exhibit lower accuracy.} All models specifically adapted for chart-related tasks perform significantly worse than general-purpose models. This may be attributed to the smaller training scale of chart-specific models compared to general-purpose models, as well as the lack of complex reasoning data in their training corpus.

\subsection{Impact of Chain of Thought Reasoning}
Multi-chart questions typically require sophisticated reasoning skills. One widely used method to enhance reasoning ability is the Chain of Thought (CoT) strategy. In this section, we test the impact of applying the CoT strategy when evaluating models on our benchmark. We explicitly define the response format in the instruction to control the model's output. Specifically, we compare two settings: original and w/o CoT. In the original setting, we encourage models to generate intermediate calculations before providing the final answer (the same setting as the results shown in Table~\ref{tab:main_result}). As to the w/o CoT setting, results are obtained by instructing models not to generate any explanations or intermediate calculation steps. Detailed instructions are provided in Appendix~\ref{appendix:prompt}. Comparison results are illustrated in  Figure~\ref{fig:CoT_&_merged}. It shows that 1) most MLLMs experience a significant performance drop in the w/o CoT setting, especially among closed-source models, and 2) half of the open-source models maintain or even improve their performance without CoT. One possible explanation is that chain-of-thought, as an emergent ability~\cite{cot}, does not result in performance gains for smaller models ($\sim 7B$ parameters). This comparison demonstrates that the MultiChartQA benchmark demands advanced reasoning capabilities, underscoring its inclusion of complex reasoning scenarios.

\subsection{Results under Chart Merging Setting}
Some MLLMs adopt limited training data with multiple images, which could negatively impact their performance in multi-image evaluation~\cite{chen2024far}. One approach to address this issue is to preprocess multiple images into one image by merging them horizontally before feeding them into the model. In this section, we evaluate the performance of different models on MultiChartQA using two settings: original multi-image input and merged image input. We merge all the charts in a set into a single composite chart by sequentially concatenating them from left to right. Performance comparison is shown in Figure~\ref{fig:CoT_&_merged}. Most MLLMs exhibit a decline in accuracy when processing these merged charts. A possible explanation is that the merged charts lose the individual semantic distinctions between separate charts, which are essential for accurate cross-chart reasoning.

\begin{table}
    \centering
    \scalebox{0.84}{
         \begin{tabular}{lccccc@{}}
        \hline
        \toprule
        \multirow{3}{*}{\textbf{Model}} & \multicolumn{2}{c}{\textbf{Cross-Chart}} &  & \multicolumn{2}{c}{\textbf{Direct}} \\
        
        \cmidrule{2-3} \cmidrule{5-6}
        
        \phantom{}  & \small
        \textbf{w/ Ref.} & \small \textbf{w/o Ref.} &  & \small \textbf{All} & \small \textbf{Spec.} \\

        \midrule
        
        Claude-3.5-Sonnet  & \textbf{63.10} & \textbf{60.69} &  &
        \textbf{82.55} & \textbf{84.43} \\
        
        GPT-4o  &  49.60 & 46.25 &  & 78.30 & \textbf{84.43}\\
        
        Gemini-1.5-Pro  &  42.29 & 38.74 &  &  69.34 & 72.17\\
        
        GPT-4o-mini  &  37.75 & 39.13 &  & 62.74 & 71.70\\ 

        MiniCPM-V2.6 & 31.69 & 30.51 &  & 64.62 & 70.28 \\

        InternVL2-26B &  31.69 & 29.72 &  & 62.26 & 70.75 \\
        
        InternVL-V1.5  &  27.27 & 27.08 &  & 54.25 & 65.09\\
        
        Qwen2-VL-7B  &  29.53 & 28.54 &  & 59.43 & 63.68\\
        
        LLaVA-OV-7B &  24.41 & 24.02 &  & 48.11 & 56.60\\
        
        MiniCPM-V2.5  &  27.08 & 26.28 & & 38.21 & 58.02\\
        
        DeepSeek-VL-7B  &  19.57 & 22.73 &  & 36.32 & 41.51\\
        
       LLaVA-V1.6-7B &  24.90 & 17.79 & & 26.42 & 35.85\\
        
        Idefics3-8B &  11.42 & 11.42 &  & 11.32 & 20.75\\      
        
        Idefics2-8B  &  8.50 & 8.10 &  & 10.38 & 22.64\\
        
        \bottomrule
        \hline
        \end{tabular}
    }
    \caption{We evaluate the performance of 14 MLLMs under two conditions: with and without chart reference, and with all charts or only specified charts provided. \textbf{Bold} values represent the best performance across all models. Due to space constraints, the experiment settings are abbreviated as follows: w/ Ref.: With Reference, w/o Ref.: Without Reference, All: All Charts, Spec.: Specified Chart Only.}
  \vspace{-2ex}
  \label{tab:Chart Reference}
\end{table}

\subsection{The Impact of Chart Reference}

We further analyze the impact of including chart references in the input questions. Chart references are phrases that specify which charts the model should focus on to facilitate information localization, such as \textit{how many members of the U.S. House are aged 50-59 the 118th Congress \textbf{in the third chart}?} We conduct two sets of experiments: \textit{Remove chart references.} We sample 254 questions from spanning parallel question answering, comparative, and sequential reasoning tasks. The questions are chosen in a way that removing chart references would not impair their answerability. We then remove the chart references from the input questions and evaluate the model's performance. According to the results in Table~\ref{tab:Chart Reference}, the absence of chart references led to a performance drop in most MLLMs. The breakdown of results by question type is detailed in Table~\ref{tab:detail_w_w/o_reference}. These findings suggest that MLLMs still face challenges with information localization in multi-chart scenarios, often relying on chart references to accurately locate the required information. \textit{Only input the specified chart.} Next, we sample 212 questions from the direct question-answering category and evaluate the model using only the specific chart that contains the required information. As illustrated in Table~\ref{tab:Chart Reference}, providing only the relevant chart leads to accuracy improvements across all models (see Table~\ref{tab:detail_chart_input} for detailed results). Notably, for relatively weaker models such as MiniCPM-Llama3-V2.5, Idefics3-8B, and Idefics2-8B, the accuracy gains are particularly substantial. The improvements are especially pronounced for relatively weaker models, such as MiniCPM-Llama3-V2.5, Idefics3-8B, and Idefics2-8B. These findings suggest that MLLMs are easily distracted by irrelevant charts and face challenges in information localization in multi-image settings. This highlights the unique difficulty posed by MultiChartQA compared to previous single-chart evaluations.


\subsection{Error Analysis}
\label{4.5:Error analysis}

We conduct a detailed error analysis on the top-performing model, \ie Claude-3.5-Sonnet, on MultiChartQA. We randomly select 60 error cases and categorize them into 3 distinct types. Case demonstration can be found in Appendix~\ref{appendix:error}.

\noindent \textbf{Perceptual errors (37\%).} This category comprises: (1) \textit{Compositional perception errors (18\%)}, where the model fails to identify key patterns, such as overall trends or the maximum and minimum values in the chart. (2) \textit{Color perception errors (9\%)}, where the model misinterprets the specified color in the legend. (3) \textit{Spatial perception errors (5\%)}, where the model struggles to locate information based on directional cues, such as "the topmost on the left." (4) \textit{OCR errors (5\%)}, where the model identifies the location of the relevant information but fails to precisely convert it into the corresponding text. This type of error indicates the significant visual perception challenges that MLLMs encounter in multi-chart scenarios.

\noindent \textbf{Reasoning errors (45\%).} In this case, the model accurately perceives the charts and extracts the relevant visual information. However, it fails in the language reasoning process required to arrive at the correct answer. These errors indicate that performing precise, step-by-step reasoning using information from multiple charts remains a significant challenge for MLLMs.

\noindent \textbf{Instruction understanding errors (18\%).} This type of error occurs when the model overlooks or misinterprets a critical constraint specified in the question. For example, if the model is asked to count identical tick values across all axes but only considers those on the vertical axes, it misinterprets which part of the chart the instruction refers to. This demonstrates a weakness in the model's ability to conjunctively process both the instructions and chart structures.

%% file: sections/conclusion.tex
\section{Conclusion}
We introduce MultiChartQA, a comprehensive benchmark for evaluating multi-chart understanding using semantically related chart sets and questions spanning multiple categories. Testing 20 MLLMs reveals a significant performance gap between closed-source and open-source models. Incorporating Chain-of-Thought reasoning notably improves accuracy. MultiChartQA highlights crucial avenues for advancing MLLMs in multi-chart reasoning tasks.

\section{Limitations}

The key limitation of our dataset lies in the cost of the collection of samples, resulting in a limited scale of MultiChartQA. This is because each question in our multi-chart sets is carefully crafted by human experts and the corresponding answers are thoroughly validated to ensure their accuracy. However, the labor-intensive nature of this process also incurs significant costs in terms of time and resources. As a result, the overall scale of our multi-chart sets is relatively limited compared to larger datasets. We leave scaling up the sample number using semi-automate methods as future work.

\section*{Acknowledgments}

This work was supported by NSF IIS-2119531, IIS-2137396, IIS-2142827, IIS-2234058, CCF-1901059, and ONR N00014-22-1-2507.
This research was supported by the International Summer Undergraduate Research Experience (iSURE) Program at the University of Notre Dame. We extend our gratitude to Zhuiri Xiao, Yuyang Tian, Zihao Xu, Sicheng Yang, Heng Wang, Tianyuan Yang, Yanyu Yao, Shuaicheng Ren, Bowen Zhou, Xinyang Yin, and Yuyang Bai for their participation in the human evaluation.

%% file: appendices/question_categories.tex
\section{Examples of All Question Categories.}

This section provides examples of questions from four distinct categories. For each subcategory, a representative example has been chosen to demonstrate our approach to question design. The distribution of questions across these subcategories is shown in Figure~\ref{fig:question_statistics}.

\begin{figure*}[h!]
    \includegraphics[width=\textwidth]{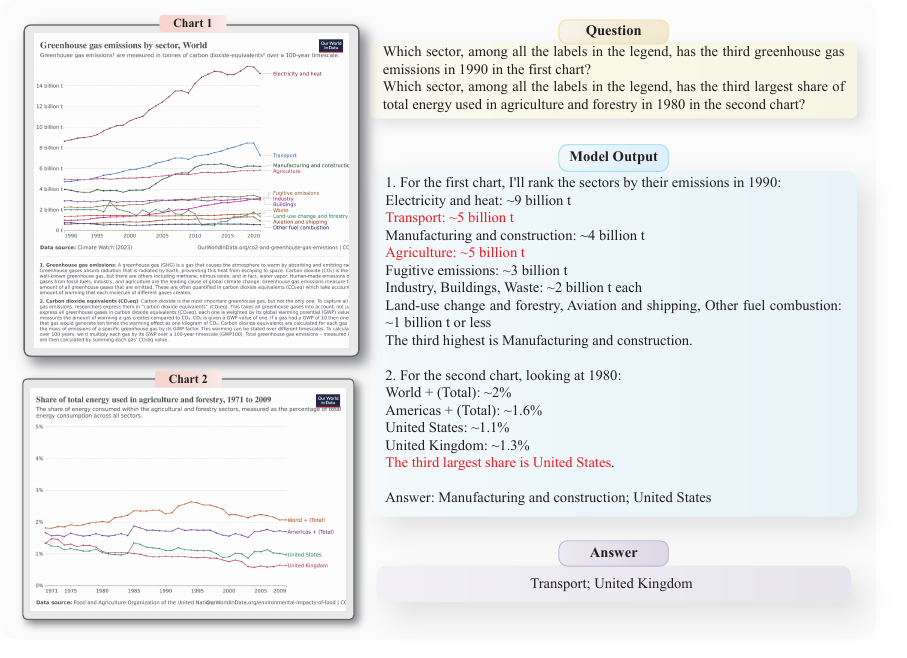}
    \caption{An example of error analysis for a chart set is illustrated, with the corresponding question, Claude-3.5-Sonnet output, and correct answer shown on the right.}
    \label{fig:error}
\end{figure*}

\begin{table*}[h]
  \centering
    \scalebox{1.0}{
    \begin{tabular}{p{3.8cm}p{2cm}p{8.2cm}}
    
    \hline
    \toprule
    \textbf{Main Category} & \textbf{Subcategory} & \textbf{Example} \\
    \midrule
     Direct Questions & Structure & How many subplots are there in the third chart? \\
     \cmidrule{2-3}
    & Content & Which age group has the lowest usage rate of ChatGPT in the first chart? \\
    \midrule
    Parallel Questions & Structure & 1. What are the names of the labels in the legend of the third chart? (from top to bottom, then left to right) \newline
    2. What is the difference between consecutive numerical tick values on the horizontal axis in the second chart? \\
    \cmidrule{2-3}
    &  Mixed & 1. What is the peak number of whales killed in a single year according to the first chart? \newline
    2. What does the horizontal axis of the second chart represent? \newline
    A. The years from 1890 to 2001. \newline
    B. The decline in global whale populations over time. \newline
    C. The percentage change in whale populations from 1890 to 2001. \newline
    D. The total whale populations in the pre-whaling period versus 2001.\\
    \cmidrule{2-3}
    & Content & 1. Which conference's annual attendance reaches its maximum in the year of 2020 in the second chart? \newline
    2. In which year the Minority stake investment reaches its maximum among the all years listed on the horizontal axis in the third chart?\\
    \midrule
    Comparative Reasoning & Structure & How many of the same tick values are there on the vertical axis in the second chart and the first chart?\\
    \cmidrule{2-3}
    & Content &  In the country with the light blue color on the left, in how many charts is the label 'Costs of Goods and Services' higher than the label 'Compensation of Employees'?\\
    \midrule
    Sequential Reasoning & Content & Firstly identify the age group whose percentage of people who have a parent 65+ and have a child younger than 18 is the largest among all the age groups in the first chart. \newline
    Then in which issue the percentage of people in this age group is also the largest among all the age groups in the second chart?\\
    
    \bottomrule
    \hline
    \end{tabular}
    }
    \caption{Illustrations of example questions from four categories and the subcategories.}
    \label{tab:examples_of_Question_category}
\end{table*}

\section{Error Analysis Example}
\label{appendix:error}
To provide a clear demonstration, Figure~\ref{fig:error} presents a question from the MultichartQA dataset along with the input given to Claude-3.5-Sonnet. This question belongs to the parallel question category, comprising two subquestions. In answer to the first subquestion, due to insufficient perception accuracy, the model struggles to distinguish the difference in the values of greenhouse gas emissions from transport and agriculture in 1990, as shown in Chart 1, leading to an incorrect answer. For the second subquestion, the model accurately identifies the four values representing the share of energy but fails to correctly rank these values and reason about the third largest share. These two errors highlight that MultichartQA requires both perceptual and reasoning abilities to answer questions correctly.

\section{Evaluation Results Without CoT.}

In this section, we present the detailed evaluation results without Chain-of-Thought (CoT) across all categories of questions. Table~\ref{tab:detail_w/o_CoT} shows the detailed performance across different question categories. Most models show a notable decline in accuracy, particularly in comparative reasoning involving structural analysis. One likely reason is the increased difficulty for MLLMs to handle numerical calculations without CoT, especially when multi-hop reasoning is needed. In this scenario, writing down the intermediate reasoning steps before arriving at the final answer can facilitate the resolution of these questions.

\begin{table*}
  \centering
  \scalebox{0.91}{
      \begin{tabular}{lrrrrrrrrrrrrrr@{}}
        \hline
        \toprule
        \multirow{3}{*}{\textbf{Model}} &
        \multirow{3}{*}{\textbf{Overall}} & \phantom{} &
        \multicolumn{2}{c}{\textbf{Direct}} & \phantom{} &
        \multicolumn{3}{c}{\textbf{Parallel}} & \phantom{} &
        \multicolumn{2}{c}{\textbf{Comparative}} & \phantom{} &
        \textbf{Seq.} & \phantom{} \\
        
        \cmidrule{4-5} \cmidrule{7-9} \cmidrule{11-12} \cmidrule{14-14}
        
        \phantom{} & \phantom{} & & \small \textbf{Struct.} & \small \textbf{Cont.} &  & \small \textbf{Struct.} & \small \textbf{Mixed} & \small \textbf{Cont.} & & \small \textbf{Struct.} & \small \textbf{Cont.} &  & \small \textbf{Cont.} \\

        \midrule
            \multicolumn{15}{c}{\textbf{Closed-source Models}} \\
        \midrule

        Claude-3.5-Sonnet  & \textbf{65.38} &  & \textbf{76.22} & \textbf{74.27} &   & \textbf{72.12} & \textbf{77.45} & \textbf{71.66} &   & 36.02 & \textbf{70.27} &   & \textbf{42.91}  \\

        GPT-4o  & 59.40 &  & 68.18 & 64.73 &   & 61.41 & 67.94 & 66.13 &   & \textbf{36.49} & 66.62 &   & 39.64  \\

        Gemini-1.5-Pro  & 53.81 &  & 64.69 & 61.41 &   & 55.45 & 62.41 & 55.76 &   & 32.70 & 58.65 &   & 35.64  \\

        GPT-4o-mini  & 42.89 &  & 59.09 & 52.70 &   & 31.21 & 42.48 & 50.00 &   & 13.27 & 54.86 &   & 25.82  \\

        \midrule
            \multicolumn{15}{c}{\textbf{Open-source Models}} \\
        \midrule

        MiniCPM-V2.6 & 40.20 &  & 51.05 & 53.53 &   & 28.89 & 38.01 & 39.17 &   & \textbf{21.33} & 58.38 &   & 16.73  \\

        InternVL2-26B & 44.38 &  & 59.79 & 53.11 &   & \textbf{42.63} & \textbf{51.63} & 49.77 &   & 10.43 & 53.78 &   & \textbf{24.73}  \\

        Qwen2-VL-7B  & \textbf{46.99} &  & \textbf{64.34} & \textbf{63.90} &   & 40.61 & 48.87 & \textbf{51.15} &   & 17.06 & \textbf{61.08} &   & 17.09  \\
        
        InternVL-V1.5  & 42.76 &  & 58.74 & 56.02 &   & 41.41 & 49.50 & 49.31 &   & 10.43 & 51.22 &   & 17.82  \\

        LLaVA-OV-7B & 36.26 &  & 51.05 & 43.98 &   & 30.30 & 35.18 & 38.25 &   & 13.27 & 52.03 &   & 13.45  \\

        MiniCPM-V2.5  & 30.63 &  & 40.56 & 26.14 &   & 31.01 & 33.97 & 32.26 &   & 10.43 & 45.27 &   & 15.64  \\

        DeepSeek-VL-7B  & 26.76 &  & 32.87 & 30.29 &   & 23.33 & 26.45 & 27.65 &   & 12.80 & 40.68 &   & 10.91  \\

        Eagle-X5-13B & 26.73 &  & 34.27 & 38.17 &   & 20.00 & 25.60 & 25.12 &   & 10.43 & 37.03 &   & 13.82  \\

        LLaVA-V1.6-7B & 23.06 &  & 25.52 & 24.48 &   & 19.80 & 23.83 & 28.80 &   & 12.80 & 35.14 &   & 7.64  \\
        
        Idefics3-8B & 13.23 &  & 8.39 & 11.20 &   & 8.08 & 11.99 & 16.36 &   & 4.27 & 26.08 &   & 11.27  \\
        
        Idefics2-8B  & 9.08 &  & 4.90 & 9.96 &   & 6.77 & 8.09 & 8.06 &   & 3.79 & 17.03 &   & 9.09  \\

        \midrule
            \multicolumn{15}{c}{\textbf{Chart-Domain Specialized Models}} \\
        \midrule

        ChartGemma & \textbf{10.91} &  & \textbf{9.09} & \textbf{15.35} &   & 4.55 & 7.94 & 10.37 &   & 2.84 & \textbf{19.59} &   & \textbf{10.18}  \\

        TinyChart-3B  & 8.82 &  & 5.24 & 7.88 &   & \textbf{6.36} & \textbf{9.72} & \textbf{11.06} &   & 3.79 & 15.68 &   & 6.91  \\
        
        MatCha & 7.58 &  & 6.99 & 10.79 &   & 2.83 & 2.77 & 5.30 &   & \textbf{5.69} & 17.30 &   & 2.55  \\

        \bottomrule
        \hline
      \end{tabular}
    }
    \caption{Evaluation results on MultiChartQA without Chain of Thought. \textbf{Bold} values indicate the best performance within each category.}
  \vspace{-2ex}
  \label{tab:detail_w/o_CoT}
\end{table*}

\section{Evaluation Results with Merged Charts.}

In this section, we present the evaluation results using merged charts across all subcategories. Most models show a decline in accuracy across the majority of subcategories. Notably, the accuracy for comparative and sequential reasoning decreases more significantly than for direct and parallel questions. This could be because comparative and sequential reasoning are inherently more complex, making it more difficult for MLLMs to generalize from single-chart scenarios to multi-chart ones, especially when faced with more challenging questions.

\begin{table*}
  \centering
  \scalebox{0.91}{
      \begin{tabular}{lrrrrrrrrrrrrrr@{}}
        \hline
        \toprule
        \multirow{3}{*}{\textbf{Model}} &
        \multirow{3}{*}{\textbf{Overall}} & \phantom{} &
        \multicolumn{2}{c}{\textbf{Direct}} & \phantom{} &
        \multicolumn{3}{c}{\textbf{Parallel}} & \phantom{} &
        \multicolumn{2}{c}{\textbf{Comparative}} & \phantom{} &
        \textbf{Seq.} & \phantom{} \\
        
        \cmidrule{4-5} \cmidrule{7-9} \cmidrule{11-12} \cmidrule{14-14}
        
        \phantom{} & \phantom{} & & \small \textbf{Struct.} & \small \textbf{Cont.} &  & \small \textbf{Struct.} & \small \textbf{Mixed} & \small \textbf{Cont.} & & \small \textbf{Struct.} & \small \textbf{Cont.} &  & \small \textbf{Cont.} \\

        \midrule
            \multicolumn{15}{c}{\textbf{Closed-source Models}} \\
        \midrule

        Claude-3.5-Sonnet  & \textbf{69.26} &  & 73.78 & \textbf{71.78} &   & 68.18 & \textbf{72.84} & 71.43 &   & \textbf{65.40} & 73.11 &   & \textbf{56.00}  \\

        GPT-4o  & 67.25 &  & \textbf{77.62} & 63.90 &   & \textbf{72.53} & 71.21 & \textbf{71.66} &   & 60.19 & \textbf{73.92} &   & 45.82  \\

        Gemini-1.5-Pro  & 59.83 &  & 68.18 & 61.83 &   & 63.13 & 67.02 & 59.91 &   & 54.50 & 62.43 &   & 41.82  \\

        GPT-4o-mini & 52.44 &  & 63.99 & 58.51 &   & 52.53 & 53.69 & 50.46 &   & 41.71 & 59.59 &   & 34.18  \\
        
        \midrule
            \multicolumn{15}{c}{\textbf{Open-source Models}} \\
        \midrule

        MiniCPM-V2.6 & 34.12 &  & 38.46 & 26.56 &   & 30.81 & 34.68 & 34.79 &   & 25.59 & 48.24 &   & 24.73  \\

        InternVL2-26B & \textbf{52.18} &  & \textbf{66.43} & \textbf{58.92} &   & \textbf{53.54} & \textbf{56.24} & \textbf{49.54} &   & \textbf{33.18} & \textbf{59.86} &   & \textbf{33.45}  \\

        Qwen2-VL-7B  & 42.78 &  & 57.34 & 53.94 &   & 36.26 & 43.33 & 44.47 &   & 13.74 & 55.54 &   & 25.09  \\

        InternVL-V1.5  & 45.23 &  & 58.04 & 47.30 &   & 38.69 & 47.94 & 44.24 &   & 26.07 & 57.30 &   & 30.91  \\

        LLaVA-OV-7B & 31.47 &  & 43.01 & 31.54 &   & 25.35 & 29.01 & 31.34 &   & 10.90 & 50.68 &   & 15.27  \\

        MiniCPM-V2.5  & 25.47 &  & 31.47 & 16.60 &   & 20.81 & 24.54 & 24.65 &   & 12.80 & 43.78 &   & 16.36  \\

        DeepSeek-VL-7B  & 19.65 &  & 18.18 & 13.28 &   & 15.56 & 17.38 & 16.36 &   & 12.80 & 39.73 &   & 12.00  \\

        Eagle-X5-13B & 28.88 &  & 33.22 & 31.12 &   & 25.56 & 26.31 & 29.03 &   & 12.32 & 46.35 &   & 15.64  \\

        LLaVA-V1.6-7B & 20.37 &  & 18.18 & 15.77 &   & 14.85 & 20.57 & 19.82 &   & 12.32 & 36.35 &   & 14.91  \\
        
        Idefics3-8B & 15.00 &  & 19.93 & 14.52 &   & 12.32 & 14.75 & 10.37 &   & 3.79 & 25.81 &   & 9.82  \\  
        
        Idefics2-8B  & 13.70 &  & 13.99 & 14.11 &   & 6.67 & 8.72 & 11.75 &   & 7.11 & 24.32 &   & 13.82  \\

        \midrule
            \multicolumn{15}{c}{\textbf{Chart-Domain Specialized Models}} \\
        \midrule

        ChartGemma & 8.97 &  & 3.85 & 4.56 &   & 5.45 & \textbf{9.50} & \textbf{10.14} &   & 3.32 & 16.49 &   & \textbf{13.09}  \\

        TinyChart-3B  & \textbf{9.75} &  & \textbf{8.74} & \textbf{6.22} &   & \textbf{7.07} & 8.65 & 7.14 &   & \textbf{6.64} & \textbf{17.43} &   & 10.55  \\
        
        MatCha & 6.33 &  & 2.80 & 2.90 &   & 2.73 & 3.26 & 1.61 &   & 5.21 & 19.19 &   & 5.09  \\

        \bottomrule
        \hline
      \end{tabular}
    }
    \caption{Evaluation results on the merged-chart setting. \textbf{Bold} values indicate the best performance within each category.}
  \vspace{-2ex}
  \label{tab:detail_merged-chart}
\end{table*}

\section{Detailed Evaluation Results of Chart Reference.}

\paragraph{With or without chart references.} We present the accuracy results for all categories included in the Cross-chart Reasoning set. Most MLLMs show a slight decline in accuracy across the majority of question categories.

\begin{table*}
  \centering
  \scalebox{1.0}{
      \begin{tabular}{lcccccccccc@{}}
        \hline
        \toprule
        \multirow{3}{*}{\textbf{Model}} & 
        \multicolumn{4}{c}{\textbf{With Reference}} & \phantom{} &
        \multicolumn{4}{c}{\textbf{Without Reference}} & \phantom{} \\

        \phantom{} & \small \textbf{Overall} & \small \textbf{Para.} & \small \textbf{Comp.} & \small \textbf{Seq.} & & \small \textbf{Overall} & \small \textbf{Para.} & \small \textbf{Comp.} & \small \textbf{Seq.} \\

        \cmidrule{2-5} \cmidrule{7-10}

        \midrule
            \multicolumn{11}{c}{\textbf{Closed-source Models}} \\
        \midrule

        Claude-3.5-Sonnet & \textbf{63.10} &  \textbf{72.79} & \textbf{66.67} & \textbf{52.69} & & \textbf{60.69} & \textbf{75.74} & \textbf{60.92} & \textbf{49.46} \\

        GPT-4o  & 49.60 &  66.20 & 48.85 & 37.89 &  & 46.25 &  64.08 & 43.10 & 35.79 \\

        Gemini-1.5-Pro & 42.29 &  47.18 & 43.10 & 37.89 & & 38.74 & 50.00 & 40.80 & 28.42 \\
        
        GPT-4o-mini & 37.75 &  46.48 & 37.36 & 31.58  & & 39.13 &  50.70 & 45.98 & 24.21 \\
        
        \midrule
            \multicolumn{11}{c}{\textbf{Open-source Models}} \\
        \midrule
        
        MiniCPM-V2.6 & \textbf{31.69} &  33.80 & 38.07 & \textbf{24.21} &  & \textbf{30.51} &  34.51 & \textbf{42.05} & 16.84 \\

        InternVL2-26B & \textbf{31.69} &  \textbf{36.62} & 38.07 & 22.11 & & 29.72 &  \textbf{35.92} & 36.36 & 18.95 \\

        InternVL-V1.5  & 27.27 &  30.99 & 32.18 & 20.00 & & 27.08 &  30.99 & 30.46 & \textbf{21.05} \\

        Qwen2-VL-7B  & 29.53 &  \textbf{36.62} & \textbf{40.91} & 13.68 & & 28.54 &  \textbf{35.92} & \textbf{42.05} & 10.53 \\

        LLaVA-OV-7B & 24.41 &  33.10 & 32.39 & 10.53  & & 24.02 &  30.99 & 32.95 & 10.53 \\

        MiniCPM-V2.5 & 27.08 &  28.17 & 31.61 & 22.11 & & 26.28 &  30.99 & 29.31 & 20.00 \\

        DeepSeek-VL-7B & 19.57 &  21.13 & 30.46 & 8.42 & & 22.73 &  26.76 & 28.16 & 14.74 \\

        LLaVA-V1.6-7B & 24.90 &  30.99 & 27.59 & 17.89 & & 17.79 &  28.17 & 17.24 & 10.53 \\
    
        Idefics3-8B & 11.42 &  14.79 & 17.61 & 3.16 & & 11.42 &  14.79 & 16.48 & 4.21 \\  
        
        Idefics2-8B & 8.50 &  10.56 & 11.49 & 4.21  & & 8.10 &  10.56 & 10.34 & 4.21 \\

        \bottomrule
        \hline
      \end{tabular}
    }
    \caption{Evaluation results with and without reference. \textbf{Bold} values indicate the best performance within each category.}
  \vspace{-2ex}
  \label{tab:detail_w_w/o_reference}
\end{table*}

\paragraph{Input all or the specified Chart only.} We present accuracy results for all subcategories of the direct question set. All the MLLMs show improved accuracy across most subcategories when provided with only the specified chart.

\begin{table*}
  \centering
  \scalebox{1.0}{
      \begin{tabular}{lrrrrrrrr@{}}
        \hline
        \toprule
        \multirow{3}{*}{\textbf{Model}} & 
        \multicolumn{3}{c}{\textbf{All Charts}} & \phantom{} &
        \multicolumn{3}{c}{\textbf{Specified Chart Only}} & \phantom{} \\

        \phantom{} & \small \textbf{Overall} & \small \textbf{Struct.} & \small \textbf{Cont.} & & \small \textbf{Overall} & \small \textbf{Struct.} & \small \textbf{Cont.}  \\

        \cmidrule{2-4} \cmidrule{6-8}

        \midrule
            \multicolumn{9}{c}{\textbf{Closed-source Models}} \\
        \midrule

        Claude-3.5-Sonnet & \textbf{82.55} & \textbf{85.71} & 77.22 & & \textbf{84.43} & \textbf{87.22} & 79.75 \\

        GPT-4o  &  78.30 &  78.20 & \textbf{78.48} &  & \textbf{84.43} &  86.47 & \textbf{81.01} \\
        
        Gemini-1.5-Pro & 69.34 &  72.93 & 63.29 &  & 72.17 &  69.92 & 75.95 \\
        
        GPT-4o-mini &  62.74 &  60.90 & 65.82 &  &71.70 &  73.68 & 68.35 \\
        
        \midrule
            \multicolumn{9}{c}{\textbf{Open-source Models}} \\
        \midrule

        MiniCPM-V2.6 & \textbf{64.62} &  \textbf{66.17} & \textbf{62.03} &  &  70.28 &  \textbf{74.44} & 63.29 \\

        InternVL2-26B & 62.26 & \textbf{66.17} & 55.70 &  &  \textbf{70.75} &  72.93 & \textbf{67.09} \\

        InternVL-V1.5  & 54.25 &  57.89 & 48.10 &  & 65.09 &  63.91 & \textbf{67.09} \\

        Qwen2-VL-7B  &  59.43 &  60.15 & 58.23 &  & 63.68 &  63.16 & 64.56 \\

        LLaVA-OV-7B &  48.11 &  45.86 & 51.90 &  & 56.60 &  52.63 & 63.29 \\

        MiniCPM-V2.5 &  38.21 &  48.87 & 20.25 &  & 58.02 &  61.65 & 51.90 \\

        DeepSeek-VL-7B & 36.32 & 41.35 & 27.85 &  & 41.51 &  39.10 & 45.57 \\

        LLaVA-V1.6-7B &  26.42 &  27.07 & 25.32 &  & 35.85 &  30.08 & 45.57 \\
    
        Idefics3-8B &  11.32 &  10.53 & 12.66 &   &20.75 &  21.80 & 18.99 \\
 
        Idefics2-8B &  10.38 &  6.77 & 16.46 &  & 22.64 &  21.05 & 25.32 \\

        \bottomrule
        \hline
      \end{tabular}
    }
    \caption{Evaluation results using all charts or the specified chart only. \textbf{Bold} values indicate the best performance within each category.}
  \vspace{-2ex}
  \label{tab:detail_chart_input}
\end{table*}

\section{Instructions}
\label{appendix:prompt}
We provide details about the instructions in the prompts used in our evaluation for reproducibility. These instructions are designed to specify the desired response format of MLLMs, such as whether Chain-of-Thought (CoT) reasoning is permitted. For our Random (GPT-4o) baseline, we adopt the prompt head from Charxiv \cite{wang2024charxiv}.

\begin{table*}[t!]
  \centering
    \scalebox{0.92}{
    \begin{tabular}{p{3cm}p{11cm}}
    
    \hline
    \toprule
    \textbf{Category} & \textbf{Detail} \\
    \midrule
     With CoT & Here is a question for you to solve. \newline
    You should give the response in the following format. \newline
    "Solution: (Here you are allowed to use chain of thought and provide short explanations or intermediate calculation steps.)\newline
    Answer: (You must output the answer in the format specified in the question without giving any explanations or intermediate calculation steps.)" \\
        \midrule
        
    Without CoT  & Here is a question for you to solve. \newline
    You should avoid providing any explanations in your response and only output the answer to this question in the format as follows.  \newline
    "Answer: (You must output the answer in the format specified in the question without giving any explanations or intermediate calculation steps.)" \\
        \midrule
        
    Random (GPT-4o) & Randomly guess a reasonable answer based on the question only. \newline
    If the question asks for a number, you can randomly guess a number within a reasonable range. \newline
    If the question asks for a term, you can randomly guess a term that is relevant to the question. \newline
    You should give the response in the following format. \newline
    "Solution: (Here you are allowed to use chain of thought and provide short explanations or intermediate calculation steps.) \newline
    Answer: (You must output the answer in the format specified in the question without giving any explanations or intermediate calculation steps.)" \\
    
    \bottomrule
    \hline
    \end{tabular}
    }
    \caption{Illustrations of prompts: In specific settings, such as with CoT (Chain of Thought), the corresponding prompt is prepended to the question, and the entire string is then submitted to the MLLMs.}
    \label{tab:prompt_head}
\end{table*}

\section{Multi-chart Examples}
\label{appendix:charts}
We demonstrate four sets of charts from 4 main source websites.

\begin{figure*}[]
    \includegraphics[width=\textwidth]{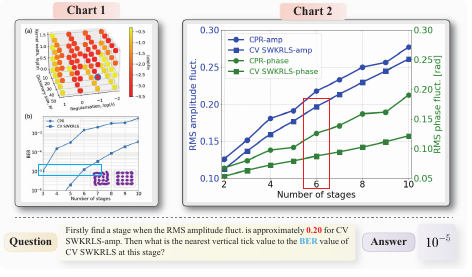}
         \caption{A chart set and the corresponding question from the ArXiv website.}
    \label{fig:example_arxiv}
\end{figure*}
\begin{figure*}[t!]
    \includegraphics[width=\textwidth]{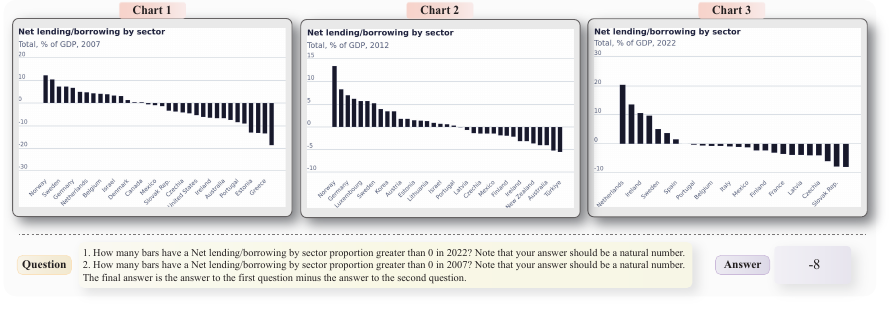}
     \caption{A chart set and the corresponding question from the OECD website.}
    \label{fig:example_oecd}
\end{figure*}
\begin{figure*}[t!]
    \includegraphics[width=\textwidth]{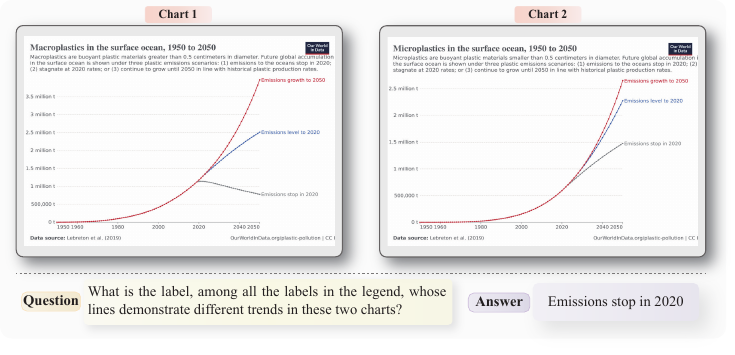}
        \caption{A chart set and the corresponding question from the Our World in Data website.}
    \label{fig:example_owid}
\end{figure*}
\begin{figure*}[t!]
    \includegraphics[width=\textwidth]{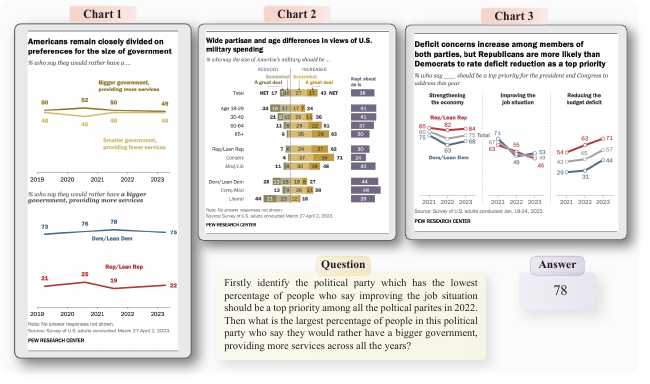}
    \caption{A chart set and the corresponding question from the Pew Research Center website.}
    \label{fig:example_pew}
\end{figure*}

\section{Chart Source}
\label{appendix:charts_source}

\begin{figure*}[t!]
    \includegraphics[width=\textwidth]{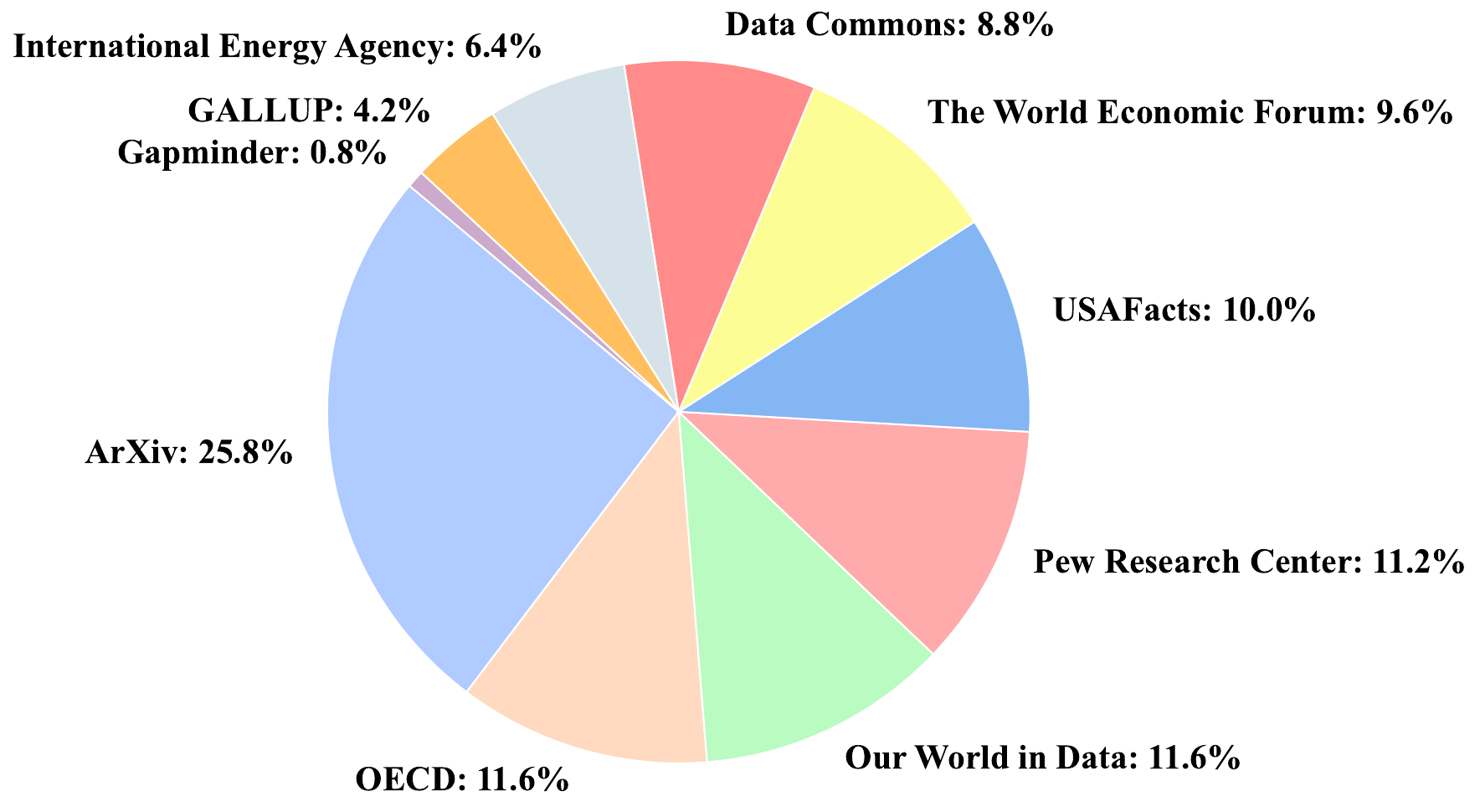}
    \caption{Illustrations of ten chart sources and their corresponding proportions.}
    \label{fig:chart_source}
\end{figure*}

We demonstrate the proportion distribution of 10 chart sources in Figure~\ref{fig:chart_source}.

%% file: acl_latex.bbl
\begin{thebibliography}{48}
\providecommand{\natexlab}[1]{#1}

\bibitem[{Agrawal et~al.(2019)Agrawal, Desai, Wang, Chen, Jain, Johnson, Batra, Parikh, Lee, and Anderson}]{agrawal2019nocaps}
Harsh Agrawal, Karan Desai, Yufei Wang, Xinlei Chen, Rishabh Jain, Mark Johnson, Dhruv Batra, Devi Parikh, Stefan Lee, and Peter Anderson. 2019.
\newblock Nocaps: Novel object captioning at scale.
\newblock In \emph{Proceedings of the IEEE/CVF international conference on computer vision}, pages 8948--8957.

\bibitem[{Anthropic(2024)}]{claude3.5-sonnet}
Anthropic. 2024.
\newblock \href {https://www-cdn.anthropic.com/fed9cc193a14b84131812372d8d5857f8f304c52/Model_Card_Claude_3_Addendum.pdf} {Claude 3.5 sonnet model card addendum}.

\bibitem[{Chen et~al.(2024)Chen, Wang, Tian, Ye, Gao, Cui, Tong, Hu, Luo, Ma et~al.}]{chen2024far}
Zhe Chen, Weiyun Wang, Hao Tian, Shenglong Ye, Zhangwei Gao, Erfei Cui, Wenwen Tong, Kongzhi Hu, Jiapeng Luo, Zheng Ma, et~al. 2024.
\newblock How far are we to gpt-4v? closing the gap to commercial multimodal models with open-source suites.
\newblock \emph{arXiv preprint arXiv:2404.16821}.

\bibitem[{Fu et~al.(2024)Fu, Hu, Li, Feng, Wang, Lin, Roth, Smith, Ma, and Krishna}]{fu2024blink}
Xingyu Fu, Yushi Hu, Bangzheng Li, Yu~Feng, Haoyu Wang, Xudong Lin, Dan Roth, Noah~A Smith, Wei-Chiu Ma, and Ranjay Krishna. 2024.
\newblock Blink: Multimodal large language models can see but not perceive.
\newblock \emph{arXiv preprint arXiv:2404.12390}.

\bibitem[{Goyal et~al.(2017)Goyal, Khot, Summers-Stay, Batra, and Parikh}]{goyal2017making}
Yash Goyal, Tejas Khot, Douglas Summers-Stay, Dhruv Batra, and Devi Parikh. 2017.
\newblock Making the v in vqa matter: Elevating the role of image understanding in visual question answering.
\newblock In \emph{Proceedings of the IEEE conference on computer vision and pattern recognition}, pages 6904--6913.

\bibitem[{Han et~al.(2023)Han, Zhang, Chen, Yang, Wang, Yu, Fu, and Zhang}]{chartllama}
Yucheng Han, Chi Zhang, Xin Chen, Xu~Yang, Zhibin Wang, Gang Yu, Bin Fu, and Hanwang Zhang. 2023.
\newblock Chartllama: {A} multimodal {LLM} for chart understanding and generation.
\newblock \emph{CoRR}, abs/2311.16483.

\bibitem[{Hoque et~al.(2022)Hoque, Kavehzadeh, and Masry}]{hoque2022chart}
Enamul Hoque, Parsa Kavehzadeh, and Ahmed Masry. 2022.
\newblock Chart question answering: State of the art and future directions.
\newblock In \emph{Computer Graphics Forum}, volume~41, pages 555--572. Wiley Online Library.

\bibitem[{Jia et~al.(2024{\natexlab{a}})Jia, Yu, Ma, Fang, Zhang, Ouyang, Zhang, Jiang, and Yu}]{leopard}
Mengzhao Jia, Wenhao Yu, Kaixin Ma, Tianqing Fang, Zhihan Zhang, Siru Ouyang, Hongming Zhang, Meng Jiang, and Dong Yu. 2024{\natexlab{a}}.
\newblock Leopard: A vision language model for text-rich multi-image tasks.
\newblock \emph{arXiv preprint arXiv:2410.01744}.

\bibitem[{Jia et~al.(2024{\natexlab{b}})Jia, Zhang, Yu, Jiao, and Jiang}]{jia2024describe}
Mengzhao Jia, Zhihan Zhang, Wenhao Yu, Fangkai Jiao, and Meng Jiang. 2024{\natexlab{b}}.
\newblock Describe-then-reason: Improving multimodal mathematical reasoning through visual comprehension training.
\newblock \emph{arXiv preprint arXiv:2404.14604}.

\bibitem[{Jiang et~al.(2024)Jiang, He, Zeng, Wei, Ku, Liu, and Chen}]{jiang2024mantis}
Dongfu Jiang, Xuan He, Huaye Zeng, Cong Wei, Max Ku, Qian Liu, and Wenhu Chen. 2024.
\newblock Mantis: Interleaved multi-image instruction tuning.
\newblock \emph{arXiv preprint arXiv:2405.01483}.

\bibitem[{Jung et~al.(2017)Jung, Kim, Song, Hwang, Lee, Kim, and Seo}]{jung2017chartsense}
Daekyoung Jung, Wonjae Kim, Hyunjoo Song, Jeong-in Hwang, Bongshin Lee, Bohyoung Kim, and Jinwook Seo. 2017.
\newblock Chartsense: Interactive data extraction from chart images.
\newblock In \emph{Proceedings of the 2017 chi conference on human factors in computing systems}, pages 6706--6717.

\bibitem[{Kafle et~al.(2018)Kafle, Price, Cohen, and Kanan}]{kafle2018dvqa}
Kushal Kafle, Brian Price, Scott Cohen, and Christopher Kanan. 2018.
\newblock Dvqa: Understanding data visualizations via question answering.
\newblock In \emph{Proceedings of the IEEE conference on computer vision and pattern recognition}, pages 5648--5656.

\bibitem[{Kahou et~al.(2017)Kahou, Michalski, Atkinson, K{\'a}d{\'a}r, Trischler, and Bengio}]{kahou2017figureqa}
Samira~Ebrahimi Kahou, Vincent Michalski, Adam Atkinson, {\'A}kos K{\'a}d{\'a}r, Adam Trischler, and Yoshua Bengio. 2017.
\newblock Figureqa: An annotated figure dataset for visual reasoning.
\newblock \emph{arXiv preprint arXiv:1710.07300}.

\bibitem[{Kazemi et~al.(2024)Kazemi, Dikkala, Anand, Devic, Dasgupta, Liu, Fatemi, Awasthi, Guo, Gollapudi et~al.}]{kazemi2024remi}
Mehran Kazemi, Nishanth Dikkala, Ankit Anand, Petar Devic, Ishita Dasgupta, Fangyu Liu, Bahare Fatemi, Pranjal Awasthi, Dee Guo, Sreenivas Gollapudi, et~al. 2024.
\newblock Remi: A dataset for reasoning with multiple images.
\newblock \emph{arXiv preprint arXiv:2406.09175}.

\bibitem[{Kojima et~al.(2022)Kojima, Gu, Reid, Matsuo, and Iwasawa}]{cot}
Takeshi Kojima, Shixiang~Shane Gu, Machel Reid, Yutaka Matsuo, and Yusuke Iwasawa. 2022.
\newblock Large language models are zero-shot reasoners.
\newblock In \emph{Advances in Neural Information Processing Systems 35: Annual Conference on Neural Information Processing Systems 2022, NeurIPS 2022, New Orleans, LA, USA, November 28 - December 9, 2022}.

\bibitem[{Lauren{\c{c}}on et~al.(2024{\natexlab{a}})Lauren{\c{c}}on, Marafioti, Sanh, and Tronchon}]{laurenccon2024building}
Hugo Lauren{\c{c}}on, Andr{\'e}s Marafioti, Victor Sanh, and L{\'e}o Tronchon. 2024{\natexlab{a}}.
\newblock Building and better understanding vision-language models: insights and future directions.
\newblock \emph{arXiv preprint arXiv:2408.12637}.

\bibitem[{Lauren{\c{c}}on et~al.(2024{\natexlab{b}})Lauren{\c{c}}on, Tronchon, Cord, and Sanh}]{laurenccon2024matters}
Hugo Lauren{\c{c}}on, L{\'e}o Tronchon, Matthieu Cord, and Victor Sanh. 2024{\natexlab{b}}.
\newblock What matters when building vision-language models?
\newblock \emph{arXiv preprint arXiv:2405.02246}.

\bibitem[{Li et~al.(2024{\natexlab{a}})Li, Zhang, Guo, Zhang, Li, Zhang, Zhang, Li, Liu, and Li}]{li2024llava}
Bo~Li, Yuanhan Zhang, Dong Guo, Renrui Zhang, Feng Li, Hao Zhang, Kaichen Zhang, Yanwei Li, Ziwei Liu, and Chunyuan Li. 2024{\natexlab{a}}.
\newblock Llava-onevision: Easy visual task transfer.
\newblock \emph{arXiv preprint arXiv:2408.03326}.

\bibitem[{Li et~al.(2024{\natexlab{b}})Li, Ge, Ge, Wang, Wang, Zhang, and Shan}]{li2024seed}
Bohao Li, Yuying Ge, Yixiao Ge, Guangzhi Wang, Rui Wang, Ruimao Zhang, and Ying Shan. 2024{\natexlab{b}}.
\newblock Seed-bench: Benchmarking multimodal large language models.
\newblock In \emph{Proceedings of the IEEE/CVF Conference on Computer Vision and Pattern Recognition}, pages 13299--13308.

\bibitem[{Liu et~al.(2022)Liu, Piccinno, Krichene, Pang, Lee, Joshi, Altun, Collier, and Eisenschlos}]{liu2022matcha}
Fangyu Liu, Francesco Piccinno, Syrine Krichene, Chenxi Pang, Kenton Lee, Mandar Joshi, Yasemin Altun, Nigel Collier, and Julian~Martin Eisenschlos. 2022.
\newblock Matcha: Enhancing visual language pretraining with math reasoning and chart derendering.
\newblock \emph{arXiv preprint arXiv:2212.09662}.

\bibitem[{Liu et~al.(2023)Liu, Wang, Yao, Chen, Song, Cho, Yacoob, and Yu}]{liu2023mmc}
Fuxiao Liu, Xiaoyang Wang, Wenlin Yao, Jianshu Chen, Kaiqiang Song, Sangwoo Cho, Yaser Yacoob, and Dong Yu. 2023.
\newblock Mmc: Advancing multimodal chart understanding with large-scale instruction tuning.
\newblock \emph{arXiv preprint arXiv:2311.10774}.

\bibitem[{Liu et~al.(2024{\natexlab{a}})Liu, Li, Li, and Lee}]{liu2024improved}
Haotian Liu, Chunyuan Li, Yuheng Li, and Yong~Jae Lee. 2024{\natexlab{a}}.
\newblock Improved baselines with visual instruction tuning.
\newblock In \emph{Proceedings of the IEEE/CVF Conference on Computer Vision and Pattern Recognition}, pages 26296--26306.

\bibitem[{Liu et~al.(2024{\natexlab{b}})Liu, Li, Li, Li, Zhang, Shen, and Lee}]{liu2024llava}
Haotian Liu, Chunyuan Li, Yuheng Li, Bo~Li, Yuanhan Zhang, Sheng Shen, and Yong~Jae Lee. 2024{\natexlab{b}}.
\newblock Llava-next: Improved reasoning, ocr, and world knowledge.

\bibitem[{Liu et~al.(2024{\natexlab{c}})Liu, Chen, Li, Fang, and Shen}]{liu2024chartthinker}
Mengsha Liu, Daoyuan Chen, Yaliang Li, Guian Fang, and Ying Shen. 2024{\natexlab{c}}.
\newblock Chartthinker: A contextual chain-of-thought approach to optimized chart summarization.
\newblock \emph{arXiv preprint arXiv:2403.11236}.

\bibitem[{Lu et~al.(2024)Lu, Liu, Zhang, Wang, Dong, Liu, Sun, Ren, Li, Sun et~al.}]{lu2024deepseek}
Haoyu Lu, Wen Liu, Bo~Zhang, Bingxuan Wang, Kai Dong, Bo~Liu, Jingxiang Sun, Tongzheng Ren, Zhuoshu Li, Yaofeng Sun, et~al. 2024.
\newblock Deepseek-vl: towards real-world vision-language understanding.
\newblock \emph{arXiv preprint arXiv:2403.05525}.

\bibitem[{Masry et~al.(2022)Masry, Long, Tan, Joty, and Hoque}]{masry2022chartqa}
Ahmed Masry, Do~Xuan Long, Jia~Qing Tan, Shafiq Joty, and Enamul Hoque. 2022.
\newblock Chartqa: A benchmark for question answering about charts with visual and logical reasoning.
\newblock \emph{arXiv preprint arXiv:2203.10244}.

\bibitem[{Masry et~al.(2024{\natexlab{a}})Masry, Thakkar, Bajaj, Kartha, Hoque, and Joty}]{chartgemma}
Ahmed Masry, Megh Thakkar, Aayush Bajaj, Aaryaman Kartha, Enamul Hoque, and Shafiq Joty. 2024{\natexlab{a}}.
\newblock Chartgemma: Visual instruction-tuning for chart reasoning in the wild.
\newblock \emph{arXiv preprint arXiv:2407.04172}.

\bibitem[{Masry et~al.(2024{\natexlab{b}})Masry, Thakkar, Bajaj, Kartha, Hoque, and Joty}]{masry2024chartgemma}
Ahmed Masry, Megh Thakkar, Aayush Bajaj, Aaryaman Kartha, Enamul Hoque, and Shafiq Joty. 2024{\natexlab{b}}.
\newblock Chartgemma: Visual instruction-tuning for chart reasoning in the wild.
\newblock \emph{arXiv preprint arXiv:2407.04172}.

\bibitem[{Methani et~al.(2020)Methani, Ganguly, Khapra, and Kumar}]{methani2020plotqa}
Nitesh Methani, Pritha Ganguly, Mitesh~M Khapra, and Pratyush Kumar. 2020.
\newblock Plotqa: Reasoning over scientific plots.
\newblock In \emph{Proceedings of the IEEE/CVF Winter Conference on Applications of Computer Vision}, pages 1527--1536.

\bibitem[{OpenAI(2024)}]{gpt4o}
OpenAI. 2024.
\newblock \href {https://openai.com/index/hello-gpt-4o/} {Hello gpt-4o}.
\newblock News announcement by OpenAI.

\bibitem[{Reid et~al.(2024)Reid, Savinov, Teplyashin, Lepikhin, Lillicrap, Alayrac, Soricut, Lazaridou, Firat, Schrittwieser et~al.}]{reid2024gemini}
Machel Reid, Nikolay Savinov, Denis Teplyashin, Dmitry Lepikhin, Timothy Lillicrap, Jean-baptiste Alayrac, Radu Soricut, Angeliki Lazaridou, Orhan Firat, Julian Schrittwieser, et~al. 2024.
\newblock Gemini 1.5: Unlocking multimodal understanding across millions of tokens of context.
\newblock \emph{arXiv preprint arXiv:2403.05530}.

\bibitem[{Saha et~al.(2018)Saha, Khapra, and Sankaranarayanan}]{saha2018towards}
Amrita Saha, Mitesh Khapra, and Karthik Sankaranarayanan. 2018.
\newblock Towards building large scale multimodal domain-aware conversation systems.
\newblock In \emph{Proceedings of the AAAI conference on artificial intelligence}, volume~32.

\bibitem[{Schwenk et~al.(2022)Schwenk, Khandelwal, Clark, Marino, and Mottaghi}]{schwenk2022okvqa}
Dustin Schwenk, Apoorv Khandelwal, Christopher Clark, Kenneth Marino, and Roozbeh Mottaghi. 2022.
\newblock A-okvqa: A benchmark for visual question answering using world knowledge.
\newblock In \emph{European conference on computer vision}, pages 146--162. Springer.

\bibitem[{Shi et~al.(2024)Shi, Liu, Wang, Liao, Radhakrishnan, Huang, Yin, Sapra, Yacoob, Shi et~al.}]{shi2024eagle}
Min Shi, Fuxiao Liu, Shihao Wang, Shijia Liao, Subhashree Radhakrishnan, De-An Huang, Hongxu Yin, Karan Sapra, Yaser Yacoob, Humphrey Shi, et~al. 2024.
\newblock Eagle: Exploring the design space for multimodal llms with mixture of encoders.
\newblock \emph{arXiv preprint arXiv:2408.15998}.

\bibitem[{Wang et~al.(2024{\natexlab{a}})Wang, Fu, Huang, Li, Liu, Liu, Ma, Xu, Zhou, Zhang et~al.}]{wang2024muirbench}
Fei Wang, Xingyu Fu, James~Y Huang, Zekun Li, Qin Liu, Xiaogeng Liu, Mingyu~Derek Ma, Nan Xu, Wenxuan Zhou, Kai Zhang, et~al. 2024{\natexlab{a}}.
\newblock Muirbench: A comprehensive benchmark for robust multi-image understanding.
\newblock \emph{arXiv preprint arXiv:2406.09411}.

\bibitem[{Wang et~al.(2023)Wang, Ge, Ding, Kankanhalli, and Shan}]{wang2023makes}
Guangzhi Wang, Yixiao Ge, Xiaohan Ding, Mohan Kankanhalli, and Ying Shan. 2023.
\newblock What makes for good visual tokenizers for large language models?
\newblock \emph{arXiv preprint arXiv:2305.12223}.

\bibitem[{Wang et~al.(2024{\natexlab{b}})Wang, Bai, Tan, Wang, Fan, Bai, Chen, Liu, Wang, Ge et~al.}]{wang2024qwen2}
Peng Wang, Shuai Bai, Sinan Tan, Shijie Wang, Zhihao Fan, Jinze Bai, Keqin Chen, Xuejing Liu, Jialin Wang, Wenbin Ge, et~al. 2024{\natexlab{b}}.
\newblock Qwen2-vl: Enhancing vision-language model's perception of the world at any resolution.
\newblock \emph{arXiv preprint arXiv:2409.12191}.

\bibitem[{Wang et~al.(2024{\natexlab{c}})Wang, Xia, He, Chen, Liu, Zhu, Liang, Wu, Liu, Malladi et~al.}]{wang2024charxiv}
Zirui Wang, Mengzhou Xia, Luxi He, Howard Chen, Yitao Liu, Richard Zhu, Kaiqu Liang, Xindi Wu, Haotian Liu, Sadhika Malladi, et~al. 2024{\natexlab{c}}.
\newblock Charxiv: Charting gaps in realistic chart understanding in multimodal llms.
\newblock \emph{arXiv preprint arXiv:2406.18521}.

\bibitem[{Xia et~al.(2024)Xia, Zhang, Ye, Yan, Liu, Zhou, Chen, Dou, Shi, Yan et~al.}]{xia2024chartx}
Renqiu Xia, Bo~Zhang, Hancheng Ye, Xiangchao Yan, Qi~Liu, Hongbin Zhou, Zijun Chen, Min Dou, Botian Shi, Junchi Yan, et~al. 2024.
\newblock Chartx \& chartvlm: A versatile benchmark and foundation model for complicated chart reasoning.
\newblock \emph{arXiv preprint arXiv:2402.12185}.

\bibitem[{Xu et~al.(2018)Xu, Bryan, Li, Zhao, and Ma}]{xu2018chart}
Shenyu Xu, Chris Bryan, Jianping~Kelvin Li, Jian Zhao, and Kwan-Liu Ma. 2018.
\newblock Chart constellations: Effective chart summarization for collaborative and multi-user analyses.
\newblock In \emph{Computer Graphics Forum}, volume~37, pages 75--86. Wiley Online Library.

\bibitem[{Xu et~al.(2023)Xu, Du, Qi, Xu, Yuan, and Guo}]{xu2023chartbench}
Zhengzhuo Xu, Sinan Du, Yiyan Qi, Chengjin Xu, Chun Yuan, and Jian Guo. 2023.
\newblock Chartbench: A benchmark for complex visual reasoning in charts.
\newblock \emph{arXiv preprint arXiv:2312.15915}.

\bibitem[{Yao et~al.(2024)Yao, Yu, Zhang, Wang, Cui, Zhu, Cai, Li, Zhao, He et~al.}]{yao2024minicpm}
Yuan Yao, Tianyu Yu, Ao~Zhang, Chongyi Wang, Junbo Cui, Hongji Zhu, Tianchi Cai, Haoyu Li, Weilin Zhao, Zhihui He, et~al. 2024.
\newblock Minicpm-v: A gpt-4v level mllm on your phone.
\newblock \emph{arXiv preprint arXiv:2408.01800}.

\bibitem[{Yin et~al.(2024)Yin, Wang, Cao, Shi, Liu, Li, Huang, Wang, Sheng, Bai et~al.}]{yin2024lamm}
Zhenfei Yin, Jiong Wang, Jianjian Cao, Zhelun Shi, Dingning Liu, Mukai Li, Xiaoshui Huang, Zhiyong Wang, Lu~Sheng, Lei Bai, et~al. 2024.
\newblock Lamm: Language-assisted multi-modal instruction-tuning dataset, framework, and benchmark.
\newblock \emph{Advances in Neural Information Processing Systems}, 36.

\bibitem[{Zhang et~al.(2024{\natexlab{a}})Zhang, Hu, Xu, Yan, Xu, Jin, Zhang, and Huang}]{tinychart}
Liang Zhang, Anwen Hu, Haiyang Xu, Ming Yan, Yichen Xu, Qin Jin, Ji~Zhang, and Fei Huang. 2024{\natexlab{a}}.
\newblock Tinychart: Efficient chart understanding with visual token merging and program-of-thoughts learning.
\newblock \emph{CoRR}, abs/2404.16635.

\bibitem[{Zhang et~al.(2024{\natexlab{b}})Zhang, Hu, Xu, Yan, Xu, Jin, Zhang, and Huang}]{zhang2024tinychart}
Liang Zhang, Anwen Hu, Haiyang Xu, Ming Yan, Yichen Xu, Qin Jin, Ji~Zhang, and Fei Huang. 2024{\natexlab{b}}.
\newblock Tinychart: Efficient chart understanding with visual token merging and program-of-thoughts learning.
\newblock \emph{arXiv preprint arXiv:2404.16635}.

\bibitem[{Zhang et~al.(2023{\natexlab{a}})Zhang, Lee, Fang, Yu, Jia, Jiang, and Barbieri}]{plug}
Zhihan Zhang, Dong-Ho Lee, Yuwei Fang, Wenhao Yu, Mengzhao Jia, Meng Jiang, and Francesco Barbieri. 2023{\natexlab{a}}.
\newblock Plug: Leveraging pivot language in cross-lingual instruction tuning.
\newblock \emph{arXiv preprint arXiv:2311.08711}.

\bibitem[{Zhang et~al.(2024{\natexlab{c}})Zhang, Liang, Yu, Yu, Jia, Yu, and Jiang}]{refaug}
Zhihan Zhang, Zhenwen Liang, Wenhao Yu, Dian Yu, Mengzhao Jia, Dong Yu, and Meng Jiang. 2024{\natexlab{c}}.
\newblock Learn beyond the answer: Training language models with reflection for mathematical reasoning.
\newblock \emph{arXiv preprint arXiv:2406.12050}.

\bibitem[{Zhang et~al.(2023{\natexlab{b}})Zhang, Wang, Yu, Xu, Iter, Zeng, Liu, Zhu, and Jiang}]{auto-instruct}
Zhihan Zhang, Shuohang Wang, Wenhao Yu, Yichong Xu, Dan Iter, Qingkai Zeng, Yang Liu, Chenguang Zhu, and Meng Jiang. 2023{\natexlab{b}}.
\newblock Auto-instruct: Automatic instruction generation and ranking for black-box language models.
\newblock \emph{arXiv preprint arXiv:2310.13127}.

\end{thebibliography}
